\documentclass[11pt]{article}
\usepackage{acl}   

\usepackage{fontspec}
\newfontfamily\dev{Shobhika-Regular.otf}[Script=Devanagari,BoldFont=Shobhika-Bold.otf]
\newcommand{\hi}[1]{{\dev #1}}                          

\usepackage{latexsym}
\usepackage{microtype}
\usepackage{graphicx}
\graphicspath{{figures/}}
\usepackage{booktabs}
\usepackage{multirow}
\usepackage{amsmath}
\usepackage{xcolor}
\usepackage[most]{tcolorbox}
\usepackage{xspace}
\usepackage{placeins}

\definecolor{cFpb}{HTML}{0B6E6E}   
\definecolor{cOk}{HTML}{1E7A33}    
\definecolor{cDrift}{HTML}{B22222} 
\newtcolorbox{promptbox}[1]{breakable, colback=black!4, colframe=black!35, boxrule=0.4pt,
  left=5pt, right=5pt, top=4pt, bottom=4pt, fonttitle=\bfseries\small, title={#1}}

\newcommand{\benchname}{RASA-Bench\xspace}

\title{Refine, Don't Reinvent: RASA-Bench for Source-Grounded Hindi Prose Refinement}

\author{
  Abdur Rahman\thanks{\ \ Corresponding author.} \quad
  Pratham Singla \quad
  Mohammad Saifullah \quad
  Vasu Sharma \\
  Pocket FM, Bengaluru, India \\
  \texttt{\{rahman.abdur, pratham.singla, mohammad.saifullah,} \\
  \texttt{vasu.sharma\}@pocketfm.com}
}

\begin{document}
\maketitle

\begin{abstract}
For a producer of Hindi creative prose, drafting is cheap and editing is not: a machine-written
episode, story, or essay reads as under-written and flat, laced with translationese, and
it sometimes quietly alters what the source says. \benchname targets that editing task
across three Hindi prose forms, measuring how well a model refines an under-written AI
draft, a task distinct from writing from scratch. Across 2{,}050 drafts and eight
refiners (16{,}400 rewrites), a four-layer protocol credits a rewrite only after it
passes a faithfulness check against its own source, then ranks the survivors by holistic
editorial preference through per-form Bradley--Terry fits whose centered coefficients
are standardized before equal-weight averaging. Staying faithful proves to be a decisive test that behaves differently by form:
gate-failure rates range from 9.1\% on short stories to 29.5\% on audio fiction, and the
drift that disqualifies a rewrite shifts from events and identity in narrative to claims
and evidence in essays. A 600-case human audit puts gate agreement at 0.925 (0.91 to 0.94
across forms), whereas subjective preference is form-sensitive and least settled for
essays, among human readers no less than the model judge. The benchmark, evaluation
protocol, and Hindi prose diagnostics are available at \url{https://abdur75648.github.io/rasa-bench}.
\end{abstract}

\section{Introduction}\label{sec:intro}
Serialized audio fiction, literary short stories, and literary essays are all produced
at scale for Hindi audiences \citep{mann2026narrativeworldbench}, and LLM-based pipelines draft each of them
by first planning an outline and then generating prose from it
\citep{storygensurvey2025,dome2024,kazi2026pauseeditablestrategyartifacts,saifullah2026narrativeworldmodelnarratologygrounded}. A draft produced this way is rarely ready to
publish: in Hindi it tends to read as under-written and flat, laced with translationese
\citep{rahman-etal-2026-calques}, and it sometimes quietly changes what the source establishes. The production reality is
therefore \emph{refinement}: the editorial work of turning a serviceable draft into
publishable prose while keeping its content intact \citep{kazi-etal-2026-adaptation}. A strong refinement lifts the
language; a weak one leaves it unimproved; and an unfaithful one rewrites the story or
the argument. Measuring this well means separating genuine prose improvement from fluent
rewriting that drifts from the source.

Evaluation has not kept pace. Creative-writing and story-generation benchmarks are well
developed for English \citep{writingbench2025,litbench2025,hanna2022,goodstorysurvey2024,storyer2022},
Indic benchmarks cover translation and broad capability
\citep{indicmteval2022,indicnlg2022,indicgenbench2024,milu2024,airavata2024,hindillmbench2025},
and preference leaderboards rank open-ended outputs against one another
\citep{mtbencharena2023}. Source-grounded \emph{refinement} has drawn far less attention
even in English \citep{rewritelm2023,coedit2023,selfrefine2023,aislop2polish2025}, and
for non-English prose, where the failure modes are sharper, it has been little
explored. We target this gap for three distinct Hindi prose forms at once.

\begin{figure*}[tbp]\centering
\includegraphics[width=0.98\textwidth]{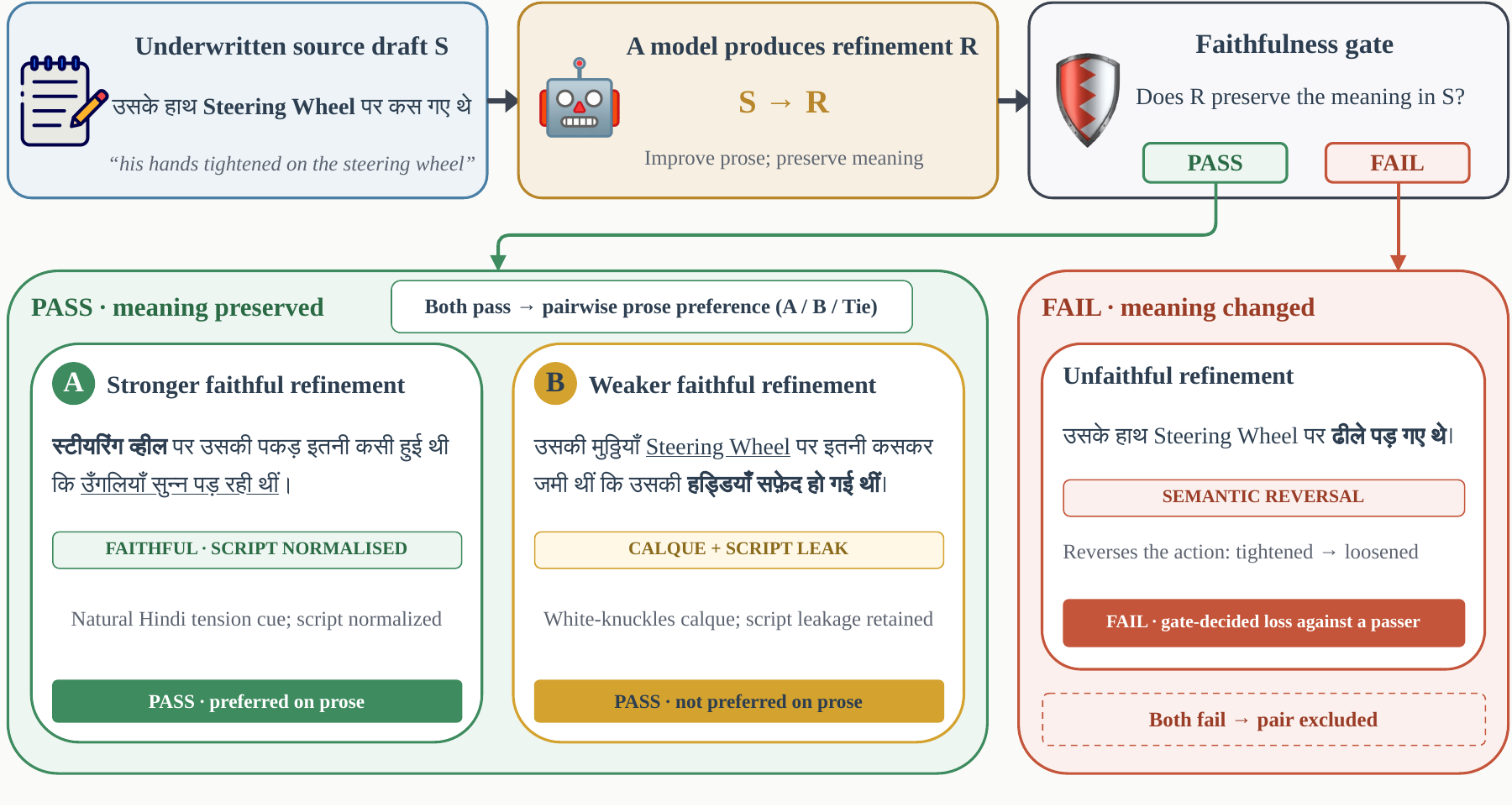}
\caption{\benchname in one object; the mechanism is shown on a narrative example but
applies to all three forms. From an under-written draft in which the narrator's hands
\emph{tighten} on the wheel, a faithfulness-violating rewrite reverses the action and is
gated out before judging; a weak but faithful rewrite imports the ``white knuckles''
calque and leaves an English-script term in place; and a strong rewrite stays faithful,
swaps the calque for a tension cue that works in Hindi, and renders the term in
Devanagari. Gate-passing rewrites then compete on holistic preference.}
\label{fig:teaser}
\end{figure*}

We present RASA-Bench (\textbf{R}efinement \textbf{A}ssessment of \textbf{S}ource-grounded
\textbf{A}I), a form-conditioned, faithfulness-gated, human-calibrated benchmark
for Hindi prose refinement across audio fiction, short stories, and literary essays. Each
model is given an under-written AI draft and must refine it; a refinement earns a quality
score only after passing a faithfulness gate against its source, and gate-passing
rewrites then compete on holistic preference (Figure~\ref{fig:teaser}). The benchmark
comprises 2{,}050 source drafts across the three forms, refined by eight models into
16{,}400 refinements and evaluated by a four-layer protocol whose primary judge is
calibrated against Hindi readers.

Two findings organize the results. First, faithfulness is itself a decisive and
\emph{form-conditioned} signal: what a rewrite must preserve differs by form, and
gate-failure rates span 9.1\% to 29.5\% across forms and a wider range across models, so
a fluent rewrite that drifts from its source is filtered before prose quality is scored.
Second, the three forms form a difficulty gradient: the objective faithfulness gate is
reliable in every form yet most discriminating on audio fiction, while subjective
preference is cleanest on short stories and genuinely low-consensus on essays, for human
readers and the automatic judge alike. The top of the leaderboard is stable across
forms; the middle reorders, and an equal-form standardized macro combines the three. Our
contributions are:

\begin{itemize}
\item \textbf{C1.} The first benchmark for source-grounded Hindi prose \emph{refinement}
across three distinct forms, audio fiction, literary short stories, and literary essays,
built from professionally edited and published human sources (\S\ref{sec:construction}).
\item \textbf{C2.} A form-conditioned, faithfulness-gated pairwise preference protocol
with equal-form standardized Bradley--Terry aggregation, human-calibrated, that separates
source-grounded refinement quality from fluent-but-unfaithful rewriting across forms
(\S\ref{sec:protocol}, \S\ref{sec:results}). It adapts faithfulness-gated filtering and
Bradley--Terry preference to this setting, reusing established preference-evaluation
components.
\item \textbf{C3.} A form-conditioned Hindi prose diagnostic signature, principally
formulaic-phrase burden and calque/translationese, that generic metrics miss and reported descriptively and kept out of the ranking, strongest in audio fiction and fainter in the
written forms (\S\ref{sec:diagnostic}).
\end{itemize}

\section{Related Work}\label{sec:related}
\textbf{Creative-writing and story evaluation.} A growing line of work benchmarks
open-ended writing quality, including WritingBench \citep{writingbench2025}, LitBench
\citep{litbench2025}, the HANNA human-annotated story corpus \citep{hanna2022}, long-horizon serialized
audio drama \citep{mann2026narrativeworldbench}, and surveys of story-quality measurement \citep{goodstorysurvey2024,storyer2022}. These
target free-form generation and report the fair-to-moderate human agreement typical of
subjective prose; HANNA, for instance, reports only fair-to-moderate inter-annotator
agreement, as we also observe, most sharply for essays. None addresses source-grounded
refinement, and literary essays in particular remain largely unstudied.

\textbf{Rewriting, refinement, and slop.} Improving a given text has been
studied as instruction-tuned editing
\citep{rewritelm2023,coedit2023}, iterative self-refinement \citep{selfrefine2023}, and
content-preserving rewriting of long-form stories across cultures
\citep{kazi-etal-2026-adaptation,kazi2026pauseeditablestrategyartifacts}; edit-based quality rewards can also select stronger
revisions of a draft \citep{aislop2polish2025}. A parallel line defines and measures ``slop'': low-quality AI
text along interpretable dimensions \citep{measuringaislop2025} and repetitive,
over-represented phrasing relative to human writing \citep{antislop2025}. We adapt the
\emph{refine-not-generate} framing to a non-English setting and add human-reviewed,
language-specific diagnostics.

\textbf{Indic and Hindi evaluation.} Indic evaluation has centred on translation,
generation, and broad capability, including IndicMT Eval \citep{indicmteval2022}, the
IndicNLG benchmark \citep{indicnlg2022}, IndicGenBench \citep{indicgenbench2024}, and
multitask suites such as MILU \citep{milu2024}, alongside Hindi-specific instruction-tuned
models and benchmarks \citep{airavata2024,hindillmbench2025}. This work targets
translation, factual, and understanding tasks; creative and literary Hindi generation,
and refinement in particular, remains largely unbenchmarked.

\begin{figure*}[tbp]\centering\includegraphics[width=0.97\textwidth]{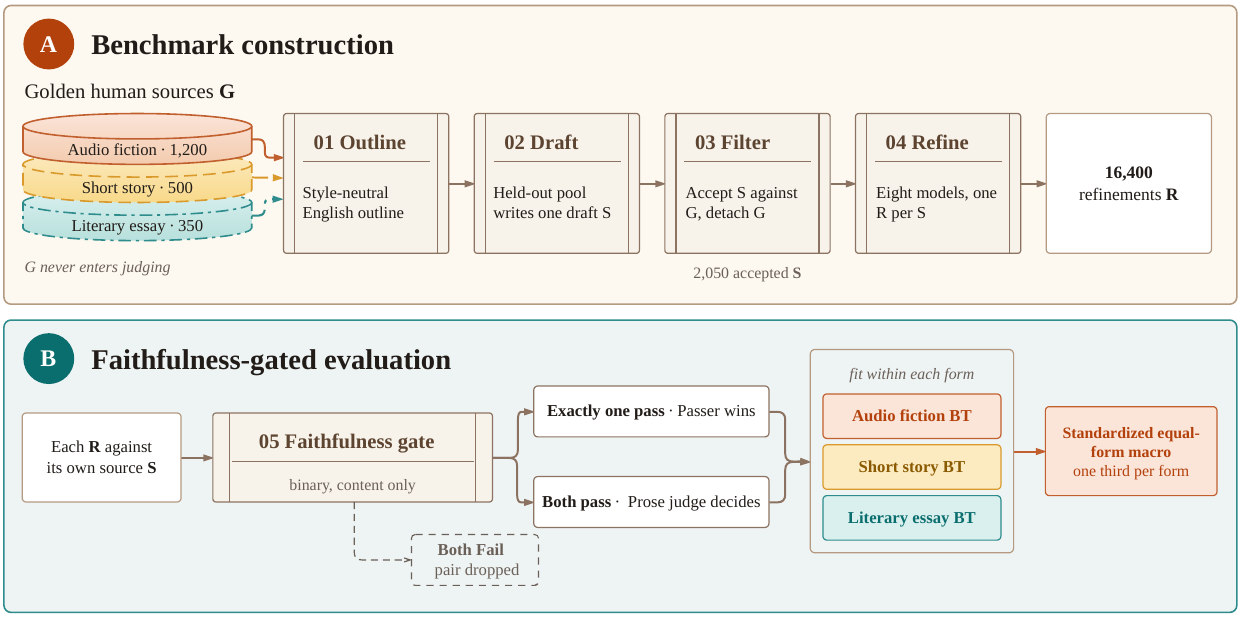}
\caption{\benchname construction and evaluation. (A) Golden human sources $G$ from
three prose forms are converted into style-neutral English outlines; held-out
generators produce under-written drafts $S$, and a form-conditioned filter accepts
each draft against $G$ before detaching $G$. Eight refiners produce 16{,}400 outputs
$R$ from 2{,}050 accepted drafts. (B) Each $R$ is checked only against its own source
draft $S$: a sole passer wins, two failures are dropped, and two passers are compared
for prose quality. Per-form Bradley--Terry coefficients are centered, scaled to unit
SD, and averaged with equal form weight. $G$ never enters leaderboard judging.}
\label{fig:construction}
\end{figure*}

\textbf{LLM-as-judge, pairwise preference, and calibration.} Pairwise comparison
underlies modern preference leaderboards \citep{mtbencharena2023}, with Bradley--Terry
aggregation of votes \citep{10.1093/biomet/39.3-4.324,chatbotarena2024}; rubric-based and model judges are now
standard \citep{geval2023,prometheus2024}. Their reliability is actively studied: judges
align with humans largely when given a correct reference \citep{nofreelabels2025}, and
design choices can leave rankings dominated by noise \citep{judgmentnoise2025}. Closest
to our protocol, WritingPreferenceBench \citep{writingprefbench2025} matches response
pairs for objective correctness so that only subjective quality is judged. We instead
impose a source-faithfulness gate and combine it with holistic pairwise preference and
human calibration, specialized to Hindi refinement across three prose forms; to our
knowledge no prior benchmark pairs a source-faithfulness gate with holistic preference
for this setting.

\section{Benchmark Construction}\label{sec:construction}

\textbf{Task.} \benchname evaluates \emph{prose refinement} across three written
Hindi forms: serialized audio fiction, literary short stories, and literary essays.
Given an under-written AI draft $S$ of one unit (an audio-fiction episode, a short
story, or an essay), a model produces a refined version $R$ that should read as
natural, register-appropriate Hindi while preserving everything the draft
establishes. This is the task a producer or editor actually faces in each form. Raw
model output is rarely publication or broadcast ready, yet it is cheap to produce;
the editorial work is to lift the language without changing what the draft says or
does. The planning is upstream and given; we therefore measure the execution,
improvement over a given draft, credited only when
the rewrite stays faithful to its source. The forms share this refinement contract but differ in what faithfulness
protects: narrative forms guard events, characters, and reveals, while essays guard
claims, evidence, and the line of reasoning.

\textbf{Construction pipeline.} Figure~\ref{fig:construction} summarizes
construction, which is identical in structure across the three forms. Each item
starts from a golden human source $G$: for audio fiction a professionally edited
episode from a private contemporary corpus, and for the written forms a published Hindi
short story or literary essay drawn from public and licensed literary collections
(Appendix~\ref{app:repro}). A style-neutral English outline, the structural plan an
upstream pipeline would emit, drives a held-out pool of generators that each execute it
into one under-written draft $S$; a form-conditioned
source-versus-gold filter accepts $S$ for core content and faithfulness (and, for
audio fiction, safety); and $G$ is then detached. Two properties hold by construction. The generator pool is held out from
the refiner roster, so no leaderboard model refines its own draft. And $G$ anchors
construction but never competes: because $R$ descends from $G$ through a lossy,
outline-mediated draft, scoring $R$ against $G$ would reward reconstruction of the
original; instead the gate checks each $R$ against its own draft $S$, and gate-passing
refinements are ranked against one another, so the benchmark measures refinement of the given draft. Reverse-engineered benchmarks
also derive each task from a finished human document, then score outputs with
checklists refined until that document satisfies every criterion
\citep{pham-etal-2026-analystbench}; \benchname detaches $G$ before evaluation. Only the form card that instantiates the outline, the draft, and the
acceptability criteria differs across forms.

\textbf{What the conversion keeps.} The filter rejects 19.0\% of candidate audio-fiction
drafts, 19.9\% of short-story drafts, and 11.2\% of essay drafts; in audio fiction it
rejects mostly for a changed or missing event, a changed character or relationship, or
a changed ending. A rejected
draft is replaced, so every retained draft has passed the check that it keeps $G$'s
core story or argument (Appendix~\ref{app:repro}). Retained drafts are shorter than $G$
(a per-item mean of 0.66 of $G$'s length in both written forms): by design the outline
withholds craft, which is what refinement restores. All eight refiners receive the identical draft, so the comparison
is controlled on that draft distribution. We take the draft in hand as the fidelity
target because an editor refines what the pipeline delivers; holding a refiner to
content it never saw would score reconstruction of the original.

\textbf{Corpus.} The accepted pool is 2{,}050 source drafts, refined by eight models
into 16{,}400 refinements: 1{,}200 audio-fiction drafts across 82 shows
(9{,}600 refinements), 500 short stories (4{,}000), and 350 literary essays
(2{,}800). Sources are drawn by stratified sampling over type and length. The
audio-fiction sources span roughly eleven predominantly contemporary genre families, led
by fantasy and cultivation (27.9\%) and family and social drama (24.4\%);
the short stories cover general literary narratives (71.6\%), fables and folk stories
(23.4\%), children's stories (4.6\%), and experimental stories; and the essays cover
cultural (59.7\%) and literary (31.7\%) reflection, personal reflection, literary
criticism, and argumentative or expository writing (full breakdown in
Appendix~\ref{app:repro}). Drafts are deliberately \emph{under-written}: a faithful
refinement lifts the prose while preserving the source's content. For audio fiction,
faithful refinements also tend to lengthen the draft, so the protocol controls for
length there (\S\ref{sec:results}); draft and refinement lengths for every form are in
Appendix~\ref{app:repro}.
\begin{table*}[tbp]\centering\small\setlength{\tabcolsep}{6pt}
\begin{tabular}{@{}rllrrrrr@{}}
\toprule
Rank & Model & Std. BT (95\% CI) & Win & Fail & J1 & J2 & J3 \\
\midrule
1 & GPT-5.5 & 1.783 [1.729, 1.831] & 0.873 & 7.7\% & \textbf{3.95} & \textbf{3.56} & \textbf{3.59} \\
2 & Claude Opus 4-6 & 1.268 [1.206, 1.322] & 0.790 & 13.9\% & 3.82 & 3.55 & 3.54 \\
3 & DeepSeek V4-Pro & 0.146 [0.102, 0.194] & 0.544 & 8.9\% & 3.55 & 3.43 & 3.40 \\
4 & Gemini 3.1 Pro & -0.033 [-0.105, 0.031] & 0.502 & 33.8\% & 3.74 & 3.46 & 3.45 \\
5 & Grok 4.3 & -0.409 [-0.460, -0.355] & 0.410 & 27.3\% & 3.61 & 3.46 & 3.43 \\
6 & Gemma 4 26B & -0.613 [-0.659, -0.566] & 0.358 & 27.8\% & 3.55 & 3.40 & 3.38 \\
7 & Sarvam 105B & -1.005 [-1.034, -0.970] & 0.268 & 10.9\% & 3.27 & 3.37 & 3.31 \\
8 & Qwen3.5 27B & -1.136 [-1.171, -1.100] & 0.239 & 18.3\% & 3.17 & 3.37 & 3.34 \\
\bottomrule
\end{tabular}
\caption{Cross-form headline leaderboard, ranked by equal-form standardized
Bradley--Terry (Std. BT). Centered per-form coefficients are scaled to unit SD before
equal-weight averaging. Win and Fail are unstandardized equal-form averages of gated
empirical win (ties count one half) and gate failure. J1--J3 are the
form-conditioned 1--5 diagnostic means (column-best in bold), descriptive within-form
reads that do not feed the ranking. Per-form tables are
in Appendix~\ref{app:perform}.}
\label{tab:leaderboard}
\end{table*}

\section{Evaluation Protocol}\label{sec:protocol}

The protocol is a form-conditioned, faithfulness-gated pairwise preference
evaluation. It adapts two established designs, faithfulness-gated filtering in the
spirit of \citet{writingprefbench2025} and holistic pairwise comparison in the
spirit of \citet{mtbencharena2023}, with Bradley--Terry ranking of pairwise votes as
in \citet{chatbotarena2024}, specialized to source-grounded Hindi refinement and
calibrated against Hindi readers. Evaluation runs in four layers. Each layer shares
one architecture across the three forms and is specialized only through a form card;
Appendix~\ref{app:prompts} lists the prompt stack and the form-conditioned content,
and gives compact specifications of the task, gate, and headline-judge prompts.

\textbf{Layer 1: faithfulness gate.} Before any quality comparison, a binary gate
judges each refinement $R$ against its source draft $S$ for content fidelity, in the
lineage of source-consistency checking for generated text \citep{summac2022}: a
rewrite that betrays the source fails, however fluent it reads. The violation
taxonomy is form-conditioned. For the narrative forms it covers changed events,
facts, timelines, reveals, dialogue meaning, character relationships, and invented content, the narrative state
that long-form fiction systems track \citep{saifullah2026narrativeworldmodelnarratologygrounded}; for essays it covers changed or dropped claims, altered or invented
evidence, and shifts in qualification or reasoning. The gate sees $S$ and runs first,
so a fluent rewrite that drifts from the source cannot win on prose. This is what
enforces \emph{refine, don't reinvent}.

\textbf{Layer 2: headline pairwise judge.} For each source, refinements that pass the
gate are compared head-to-head by a holistic judge that returns a single verdict (A, B,
or tie) rather than a scored per-dimension rubric, guided by a form-specific editorial
standard: audio-performance prose for episodes, narrative craft for short stories, and
essayistic craft for essays. A relative pairwise headline suits this
setting: once objective signals are controlled, judges align with humans only weakly on
subjective writing preference \citep{writingprefbench2025}, so we keep absolute scores
only as secondary diagnostics and rank by pairwise comparison as in public preference
leaderboards \citep{mtbencharena2023}.
Every pair is judged in both presentation orders and collapsed to one source-level
outcome to address presentation-order bias \citep{wang-etal-2024-large-language-models-fair}; any residual two-order
disagreement is collapsed to a tie (Appendix~\ref{app:robustness}). The gate and the
quality judge combine per comparison: a pair in which exactly one refinement passes the
gate is awarded to the passing one, a pair in which both fail is dropped, and a pair in
which both pass is decided by the quality judge. Collapsed outcomes feed a
Bradley--Terry model that is fit and centered \emph{separately within each form} over
that form's same-source comparisons, with ties (including the residual two-order
disagreements) entered as half a win to each model. The headline score is an
\emph{equal-form standardized Bradley--Terry macro}: each centered eight-model vector
is divided by its population SD, then each model's three standardized coefficients are
averaged. Forms thus have equal weight and equal coefficient dispersion.

\textbf{Layer 3: dimension judges.} Three 1--5 diagnostic judges decompose where
models differ, sharing a spine across forms: language and naturalness (J1),
composition and dramatization (J2), and voice and dialogue (J3), each realized
through a form-specific rubric. These are a secondary, interpretable decomposition;
because they share the headline judge's model family, their agreement with the
headline is descriptive only, and no human dimension labels were collected. The
validation of record comes from the human comparisons in Layer~2.

\textbf{Layer 4: diagnostic forensics.} A deterministic and lexical layer measures
the Hindi-specific prose signature, principally a human-reviewed formulaic-phrase
burden and a calque/translationese diagnostic. These are defined here and measured in
\S\ref{sec:diagnostic}; they are descriptive and never part of the ranking.

\begin{figure*}[tbp]\centering
\includegraphics[width=0.8\textwidth]{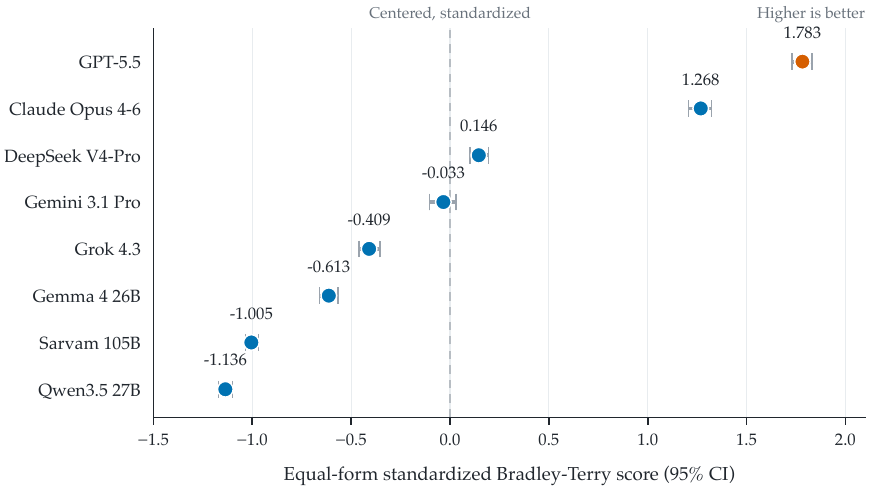}
\caption{Equal-form standardized Bradley--Terry leaderboard over 2{,}050 sources and
eight refiners. Points average three dispersion-standardized, form-specific
coefficients; bars are 95\% source-item bootstrap intervals from the full procedure.
Higher is better.}
\label{fig:leaderboard}
\end{figure*}

\textbf{Judge stack and calibration.} A single primary judge, GPT-5.5, runs the gate,
the headline comparison, and the dimensions, chosen in a pilot on Hindi and editorial
alignment \emph{independent of the leaderboard outcomes} and validated against humans
on the headline task. To probe whether one judge family decides the
ranking, two independent-family auditors (Claude Opus 4-6 and Kimi-K3) re-judged the
gate-passing audio-fiction headline comparisons involving the top model as a robustness
check (\S\ref{sec:results}); the written forms are evaluated with the single primary
judge, a scope we flag in the Limitations. Human
calibration is built into the design: Hindi readers performed the identical holistic
pairwise task in every form, and the faithfulness gate was validated against
human-labelled samples in every form (\S\ref{sec:results}). We
calibrate to ``within the human agreement band'', not to high agreement, because the
underlying preference is subjective and close-call heavy; this is the criterion MT-Bench
used to validate LLM judges, agreement with humans at the level humans reach with one
another \citep{mtbencharena2023}. Finally, because the
comparisons within a form are not independent (every model pair is judged on the same
sources), we report ranking uncertainty as Bradley--Terry scores with 95\%
\emph{source-item} bootstrap confidence intervals (1{,}000 replicates, resampling whole sources independently within each form). Each replicate repeats
the full fit, center, scale, and average procedure; we report percentile intervals.

\section{Results}\label{sec:results}

\subsection{Cross-form headline leaderboard}
On the equal-form standardized Bradley--Terry macro, adjacent models have
non-overlapping marginal 95\% source-bootstrap intervals
(Figure~\ref{fig:leaderboard}, Table~\ref{tab:leaderboard}); some per-form middle pairs
overlap (Appendix~\ref{app:perform}). GPT-5.5 leads,
followed by Claude Opus 4-6; these two are first and second in every individual form.
DeepSeek V4-Pro is a clear third on the macro, ahead of a tightly spaced middle
(Gemini 3.1 Pro, Grok 4.3, Gemma 4 26B) and a trailing pair (Sarvam 105B, Qwen3.5
27B). The middle of the table is where the forms disagree: DeepSeek, Gemini, and Grok
trade places across audio fiction, short stories, and essays (per-form leaderboards in
Appendix~\ref{app:perform}, Table~\ref{tab:perform}); the standardized macro summarizes
those form-specific swaps in one cross-form order. The dimension judges give an
interpretable, secondary read: their means lead with GPT-5.5 and Claude Opus 4-6 and
trail with the two lowest models, agreeing with the headline at the extremes, while
naturalness (J1) spreads more than composition (J2) or voice (J3).
\subsection{The gate matters}
\begin{figure}[tbp]\centering
\includegraphics[width=\linewidth]{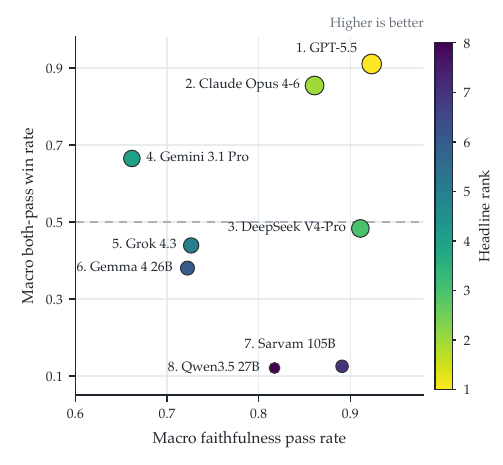}
\caption{The faithfulness-prose frontier: each model's faithfulness gate pass rate
(horizontal) against its prose quality among comparisons where both rewrites passed the
gate (vertical). GPT-5.5 and Claude Opus 4-6 lead on both axes; Gemini 3.1 Pro pairs a
lower pass rate with strong prose-only quality, while DeepSeek V4-Pro and Sarvam 105B stay
faithful with weaker prose.}
\label{fig:tradeoff}
\end{figure}

Faithfulness is one of the benchmark's decisive signals, and it behaves differently
across forms. The output-level gate-failure rate is 29.5\% for audio fiction, 9.1\%
for short stories, and 17.1\% for essays; the equal-form macro rate is 18.6\%, and the
corpus-weighted rate is 22.4\% (3{,}677 of 16{,}400 refinements). Per model the failure
rate spans a wide range and, within a model, varies by form
(Appendix~\ref{app:failfamily}, Figure~\ref{fig:failrate}): the model that drifts on nearly three in five
audio-fiction drafts (Gemini 3.1 Pro, 57.5\%) fails far less on its essays (29.4\%) and
its stories (14.6\%), so faithfulness is a joint property of both the model and the form.
Under the refinement task a faithful but plain rewrite is preferable to a
fluent one that changes the source, so every comparison between a gate-passing and a
gate-failing output is decided on fidelity before prose quality enters.
Figure~\ref{fig:tradeoff} visualizes this faithfulness--prose trade-off across models.
The headline combines each model's pass rate and its prose quality among faithful
outputs: GPT-5.5 and Claude Opus 4-6 stay
strong on both axes, while
Gemini 3.1 Pro pairs a lower pass rate with strong prose-only quality. The effect on the
ranking is real: restricting to comparisons where both rewrites passed the gate lifts
Gemini to third among faithful outputs even though faithfulness pulls its headline rank
to fourth (Appendix~\ref{app:robustness}).

The gate also labels every failure with content-violation tags, and because the
taxonomy is form-conditioned, so are the dominant failure modes
(Table~\ref{tab:vuln}). Audio-fiction failures are led by fact, object, or timeline
drift (40.9\% of failed outputs); short-story failures by relationship, identity, or
agency shifts (39.2\%); and essay failures by changed claims, inquiries, or
qualifications (65.1\%). Narrative forms most often betray a source by changing what
happens or who did it, whereas essays most often betray it by changing what is argued.
We keep these categories separate by form, since their failure types do not match
across forms. Appendix~\ref{app:diag} (Table~\ref{tab:driftexamples}) gives
human-confirmed minimal-pair examples of this drift, with the source
and rewritten spans shown side by side.

\begin{table}[tbp]\centering\small
\begin{tabular}{@{}l p{4.4cm} r@{}}
\toprule
Form & Dominant failure family & Share \\
\midrule
Audio fiction & Fact, object, or timeline drift & 40.9\% \\
Short story & Relationship, identity, or agency & 39.2\% \\
Literary essay & Claim, inquiry, or qualification & 65.1\% \\
\bottomrule
\end{tabular}
\caption{Dominant faithfulness vulnerability by prose form; shares are among that form's
failed outputs. Taxonomies are form-conditioned; complete distributions appear in
Appendix Table~\ref{tab:failfamily}.}
\label{tab:vuln}
\end{table}

\subsection{Robustness and calibration}
Length matching preserves the top two models in all three forms and the complete ranking
in short stories and essays; only lower audio-fiction ranks shift, where matching
also changes which opponents those models meet (Appendix~\ref{app:robustness},
Table~\ref{tab:lengthcontrol}, Figure~\ref{fig:length}, and
Table~\ref{tab:lengthopp}). The ranking is also
robust to generator-family kinship: four refiners share a sibling in the held-out
generator pool, yet recomputing each form's leaderboard with every sibling-matched
comparison removed preserves the order exactly in all three forms (Spearman $\rho=1.0$;
Appendix~\ref{app:kinship}).

\textbf{Ablations.} Two ablations test the design (Appendix~\ref{app:ablations}). For
the \emph{gate}, we prose-judged the 22{,}214 pairs the gate had decided or dropped, so
the judge covers all 57{,}400 same-source pairs, and graded the gate by its only
severity signal, the number of violation tags. Applying the published threshold
reproduces the published leaderboard exactly; at two tags the order is unchanged, and at
three tags and with no gate at all, only Gemini 3.1 Pro and DeepSeek V4-Pro swap third
and fourth (Spearman 0.976). Without the gate, nothing checks a rewrite against its draft,
and the model that gains is the one the gate fails most often. For an \emph{absolute anchor}, we judged each draft against its eight refinements
(16{,}400 pairs): six refiners are preferred to their own draft in every form, Sarvam
105B is indistinguishable from it, and Qwen3.5 27B falls below it on audio fiction.

Human calibration places the judge inside the band of reader agreement, and the two
layers behave differently. The objective gate is reliable in every form: over a
stratified 600-case audit (200 cases per form) it agrees with human labels on 555
(0.925, [0.901, 0.943]), holding per form at 0.930 for audio, 0.935 for stories, and
0.910 for essays. Per-form false-fail and false-pass
rates are low; they are reported with their conditioning in Appendix~\ref{app:human},
together with agreement per refiner (0.873 to 0.970, all intervals overlapping) and per
written failure family (0.879 to 0.973). Subjective preference is more
form-sensitive. On the holistic task the headline judge agrees with individual readers
on 284 of 540 labels (0.526), against an inter-human pairwise agreement of 0.472
(189/400); and it tracks the held-out reader majority at 0.524 (33/63), matching the
0.529 (83/157) at which an individual reader agrees with the leave-one-out majority of
the others. Reader agreement is itself low, as expected for subjective close-call
creative-prose preference \citep{hanna2022,judgmentnoise2025}, and it is lowest for
essays, for humans and the automatic judge alike, where the judge's agreement matches the human level (Appendix~\ref{app:human}, Table~\ref{tab:calib});
the leaderboard therefore rests on aggregate leaderboard stability, not per-item
agreement.

\begin{figure*}[tbp]\centering
\includegraphics[width=0.8\textwidth]{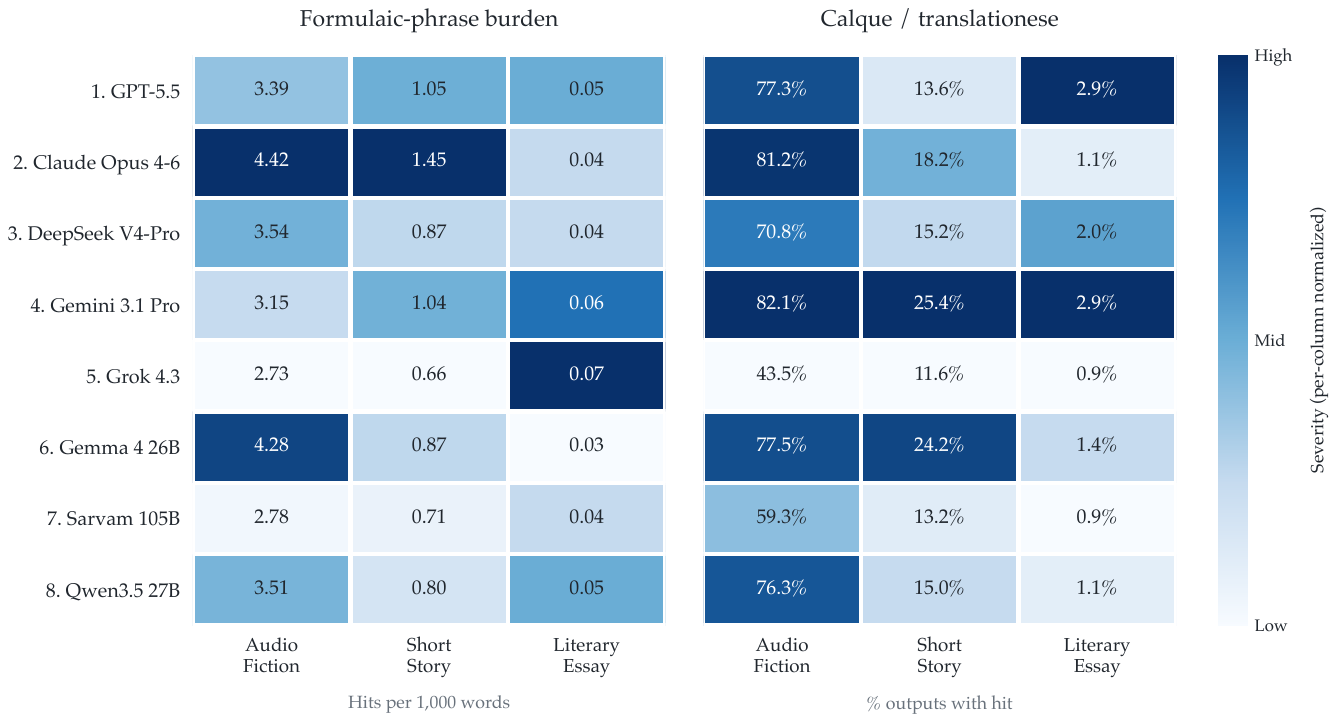}
\caption{The Hindi diagnostic signature by model and form, ordered by leaderboard rank:
formulaic-phrase burden and calque incidence. The signature is strongest in audio fiction
and progressively fainter in short stories and essays; within-form burden does not track
quality, with the strongest models carrying the most.}
\label{fig:signature}
\end{figure*}

\textbf{GPT-5.5 as judge and contestant.} GPT-5.5 is both the primary judge and the
top-ranked model, so we check each stage for self-preference. At the gate, human
labellers overturned 2 of the 37 audited GPT-5.5 outputs the gate passed (0.054,
against 0.076 across the audit), and on a re-gated audio sample our gate passed 0.856 of
GPT-5.5's outputs where an independent-family auditor passed 0.981, so the primary gate
is the stricter of the two on the top model (Appendix~\ref{app:dims}). In preference,
two independent families, Claude Opus 4-6 and Kimi-K3, re-judge the gate-passing
audio-fiction comparisons involving GPT-5.5, agree with the primary judge on 0.902 and
0.866 of them, and refitting on either preserves the full order with GPT-5.5 first
(Spearman $\rho=1.0$). The 540 human preference labels, which involve no judge, also
place GPT-5.5 first; that sample is sized to measure agreement and is too small to rank
the other models. With
no gate at all, GPT-5.5 still ranks first. The written forms rely on the single primary
judge, which we note in the Limitations.

\section{The Hindi Prose Diagnostic Signature}\label{sec:diagnostic}
Beyond the ranking, \benchname surfaces a Hindi prose signature that generic,
language-agnostic metrics miss. We report two human-reviewed diagnostics from a common
reviewed registry: a \emph{formulaic-phrase burden} (FPB), a
reviewed lexicon of stock Hindi phrasings (dialogue tags, body-reaction beats,
atmosphere and silence padding, and essay rhetorical templates) inspired by
slop-forensics work \citep{measuringaislop2025,aislop2polish2025,antislop2025}, and a
\emph{calque/translationese} diagnostic for English idioms rendered word-for-word into
Hindi. The formulaic phrasings are stock constructions that models over-produce
relative to human writing \citep{antislop2025}, so a high burden marks an analogue of
AI ``slop'' while grammaticality stays intact. Calque is a documented artifact of
multilingual generation, where source-language fingerprints surface in the target
\citep{translationesenmt2020,translationesedialects2011,translationeseunsup2015,rahman-etal-2026-calques}; we
adapt it into a reviewed Hindi diagnostic. The 120 calque surfaces are shared across all
three forms, while the formulaic lexicon has 272 audio-fiction and 274 written-form
surfaces; both layers are descriptive, and the full lexicons are in
Appendix~\ref{app:diag}.
\begin{table}[tbp]\centering\small
\renewcommand{\arraystretch}{1.2}
\begin{tabular}{@{}p{2.3cm} p{2.4cm} p{2.1cm}@{}}
\toprule
Calque (as written) & $\approx$ English idiom & Natural Hindi \\
\midrule
\hi{आवाज़ ठंडी} & ``cold voice'' & \hi{रूखी आवाज़} \\
\hi{मुस्कान आँखों तक नहीं पहुँची} & ``smile didn't reach the eyes'' & \hi{बनावटी मुस्कान} \\
\hi{रीढ़ में सिहरन दौड़ गई} & ``a shiver ran down the spine'' & \hi{रोंगटे खड़े हो गए} \\
\hi{हवा में तनाव} & ``tension in the air'' & \hi{माहौल में तनाव} \\
\bottomrule
\end{tabular}
\caption{Representative calques (real model outputs) with the natural Hindi a native
writer would use. All three forms are scored against the identical 120-surface calque
list.}
\label{tab:calques}
\end{table}

The signature is strongest in audio fiction and progressively fainter in the written
forms (Figure~\ref{fig:signature}; full per-form values in Appendix
Table~\ref{tab:lexical}): the reviewed
phrasings simply occur less often in stories and least often in essays, which does not
establish the absence of
other, form-specific formulaic language. Within a form the burden does not track
quality; if anything the stronger, more dramatized models carry more of it. In audio
fiction, formulaic burden per thousand words is 3.39 for the top model GPT-5.5 but 4.42
for the second-ranked Claude Opus 4-6 and only 2.73 for Grok 4.3, and calque appears in
82.1\% of Gemini's audio outputs but 43.5\% of Grok's, so a low burden is no sign of
better refinement. For an audio-fiction reference point, the same reviewed lexicon
appears in the human corpus at only 0.17 distinct hits per thousand words, well below
every model's audio burden. Calque is the most
legible to a native ear (Table~\ref{tab:calques}): the English ``smile didn't
reach the eyes'', traced literally into Hindi, reads as machine-made where a native
writer would use a single idiomatic phrase. Table~\ref{tab:calques} pairs four such
surfaces, each taken from a real model output, with the idiomatic Hindi a native writer
would reach for instead. Script leakage is rare: all eight models
keep a mean Latin-character fraction below 0.3\%. Appendix~\ref{app:diag}
(Table~\ref{tab:fpbexamples}) gives representative examples of these formulaic-phrase
families, with the recurring phrase highlighted.

\section{Conclusion}\label{sec:conclusion}
Refining a draft is a different demand from writing one, and it is the demand production
places on a model across forms. \benchname measures it for three kinds of Hindi prose by
asking which rewrite is stronger and crediting only the rewrites that keep faith with the
source. The results show a stable top, a form-sensitive middle, and a faithfulness signal
that is decisive and form-conditioned: what a rewrite must preserve, and how hard it is to
preserve, depends on whether it tells a story or makes an argument. The same
gated-preference design could extend to other Hindi forms and other Indic languages. The lesson for a refiner is plain: refine, don't reinvent.

\section*{Limitations}
A few boundaries frame these results. The underlying preference is subjective and
close-call heavy: readers agree only modestly, most weakly on essays (Fleiss
$\kappa=0.012$ on the 20 shared essay comparisons rated by three readers; the essay
leaderboard itself is fit on 9{,}800 pairs), so any single label is noisy and the ranking earns its reliability from
the volume of aggregated comparisons \citep{hanna2022,judgmentnoise2025}. A single model
family (GPT-5.5) serves as the primary judge across all three layers
\citep{nofreelabels2025}; \S\ref{sec:results} checks this at each stage. Our independent-family check is deliberately targeted and
correspondingly limited: two external families re-judge the top model's gate-passing
audio-fiction comparisons and leave the ranking unchanged, but the check does not extend
to the full leaderboard or to the short-story and essay forms, which rest on the single
judge; a complete cross-family, all-form audit remains future work. The faithfulness gate is
validated against a 600-case human sample, enough to show that it tracks human judgment
across forms and models, though short of an exhaustive per-model calibration, and the
construction filter that screens each draft against $G$ has no human audit of its own. The dimension judges are uncalibrated,
and their absolute scores are not comparable across forms. \benchname ranks refinements
against one another, each gated against its own draft $S$ (\S\ref{sec:construction}).
Because the short-story and essay sources are publicly available, they may appear in
model pretraining, a contamination risk the private audio corpus does not carry; the task nonetheless scores refinement of a separately generated draft $S$, and the
gate fails content that a refiner adds from memory of the source. Finally, it covers three Hindi prose forms
and leaves other forms, other Indic languages, and calibrated per-dimension scoring beyond
its present reach.

\section*{Ethics Statement}
\benchname draws on two kinds of source. The audio-fiction sources are a private corpus of
professionally edited episodes, used only to seed construction and ground the faithfulness
filter; they are never published or displayed and never enter leaderboard judging (the
construction models read them only during outline extraction and gating). The short-story
and essay sources are drawn from Hindi Wikisource \citep{hindiwikisource}, the BHAAV corpus
\citep{kumar2019bhaav}, and Hindi Discourse \citep{dhanwal2020discourse};
rights and release are handled at the item level, and some source material is restricted,
so release decisions are separate from our permission to process the texts for this
research. All released and displayed text, including every example in this paper, is
model-generated, and we do not disclose private-corpus identifying details. Human
validation was performed by native Hindi speakers, paid members of our organization who
gave informed consent and judged blind to model identity: six readers for audio-fiction
preference, three per written preference panel, and three labellers for the second
round of the gate audit (Appendix~\ref{app:human}); individual annotations, reader
identities, and demographics are withheld. The released drafts and refinements, prompts,
configurations, judge stack, evaluation code, and aggregate results for all three forms
have cleared a content, privacy, and licensing review. Only the golden human sources are private: they are used only in
construction, are not distributed, and are not needed to reproduce any result. Audio
fiction carries one third of the equal-form headline; the other two thirds rest on
public sources. The evaluation relies on LLM judges for a
subjective task alongside a human validation set, and we report agreement levels
transparently. We see the broader impact as positive: better tooling and evaluation for
refining Hindi prose serves a large audience in an underrepresented language while keeping
a human editor in the loop.

\bibliography{refs}

@article{writingprefbench2025,
  title   = {Beyond Correctness: Evaluating Subjective Writing Preferences Across Cultures},
  author  = {Ying, Shuangshuang and Li, Yunwen and Qu, Xingwei and Li, Xin and Jin, Sheng and Liu, Minghao and Wen, Zhoufutu and Du, Xeron and Zheng, Tianyu and Zhang, Yichi and Ni, Letian and Cheng, Yuyang and Chen, Qiguang and Ding, Jingzhe and Long, Shengda and Zhou, Wangchunshu and Feng, Jiazhan and Zhong, Wanjun and Qin, Libo and Zhang, Ge and Huang, Wenhao and Che, Wanxiang and Lin, Chenghua},
  journal = {arXiv preprint arXiv:2510.14616},
  year    = {2025},
  url     = {https://arxiv.org/abs/2510.14616}
}

@article{mtbencharena2023,
  title   = {Judging {LLM}-as-a-Judge with {MT-Bench} and {Chatbot Arena}},
  author  = {Zheng, Lianmin and Chiang, Wei-Lin and Sheng, Ying and Zhuang, Siyuan and Wu, Zhanghao and Zhuang, Yonghao and Lin, Zi and Li, Zhuohan and Li, Dacheng and Xing, Eric P. and Zhang, Hao and Gonzalez, Joseph E. and Stoica, Ion},
  journal = {arXiv preprint arXiv:2306.05685},
  year    = {2023},
  url     = {https://arxiv.org/abs/2306.05685}
}

@inproceedings{rewritelm2023,
  title     = {{RewriteLM}: An Instruction-Tuned Large Language Model for Text Rewriting},
  author    = {Shu, Lei and Luo, Liangchen and Hoskere, Jayakumar and Zhu, Yun and Liu, Yinxiao and Tong, Simon and Chen, Jindong and Meng, Lei},
  booktitle = {Proceedings of the AAAI Conference on Artificial Intelligence},
  volume    = {38},
  number    = {17},
  pages     = {18970--18980},
  year      = {2024},
  doi       = {10.1609/aaai.v38i17.29863},
  url       = {https://ojs.aaai.org/index.php/AAAI/article/view/29863}
}

@article{aislop2polish2025,
  title   = {{AI}-Slop to {AI}-Polish? Aligning Language Models through Edit-Based Writing Rewards and Test-time Computation},
  author  = {Chakrabarty, Tuhin and Laban, Philippe and Wu, Chien-Sheng},
  journal = {arXiv preprint arXiv:2504.07532},
  year    = {2025},
  url     = {https://arxiv.org/abs/2504.07532}
}

@inproceedings{indicmteval2022,
  title     = {{IndicMT Eval}: A Dataset to Meta-Evaluate Machine Translation Metrics for {Indian} Languages},
  author    = {Sai B, Ananya and Dixit, Tanay and Nagarajan, Vignesh and Kunchukuttan, Anoop and Kumar, Pratyush and Khapra, Mitesh M. and Dabre, Raj},
  booktitle = {Proceedings of the 61st Annual Meeting of the Association for Computational Linguistics (Volume 1: Long Papers)},
  year      = {2023},
  pages     = {14210--14228},
  publisher = {Association for Computational Linguistics},
  doi       = {10.18653/v1/2023.acl-long.795},
  url       = {https://aclanthology.org/2023.acl-long.795/}
}

@inproceedings{indicgenbench2024,
  title     = {{IndicGenBench}: A Multilingual Benchmark to Evaluate Generation Capabilities of {LLM}s on {Indic} Languages},
  author    = {Singh, Harman and Gupta, Nitish and Bharadwaj, Shikhar and Tewari, Dinesh and Talukdar, Partha},
  booktitle = {Proceedings of the 62nd Annual Meeting of the Association for Computational Linguistics (Volume 1: Long Papers)},
  pages     = {11047--11073},
  year      = {2024}
}

@inproceedings{hindillmbench2025,
  title     = {Benchmarking {Hindi} {LLM}s: A New Suite of Datasets and a Comparative Analysis},
  author    = {Kamath, Anusha and Singla, Kanishk and Paul, Rakesh and Joshi, Raviraj Bhuminand and Vaidya, Utkarsh and Chauhan, Sanjay Singh and Wartikar, Niranjan},
  booktitle = {Proceedings of the 1st Workshop on Benchmarks, Harmonization, Annotation, and Standardization for Human-Centric AI in Indian Languages (BHASHA 2025)},
  pages     = {52--68},
  year      = {2025}
}

@inproceedings{writingbench2025,
  title     = {{WritingBench}: A Comprehensive Benchmark for Generative Writing},
  author    = {Wu, Yuning and Mei, Jiahao and Yan, Ming and Li, Chenliang and Lai, Shaopeng and Ren, Yuran and Wang, Zijia and Zhang, Ji and Wu, Mengyue and Jin, Qin and Huang, Fei},
  booktitle = {Advances in Neural Information Processing Systems},
  volume    = {38},
  year      = {2025},
  doi       = {10.52202/085713-1737},
  url       = {https://proceedings.neurips.cc/paper_files/paper/2025/hash/4aedf0cba303537fcb6cf948bb41b2df-Abstract-Datasets_and_Benchmarks_Track.html}
}

@inproceedings{litbench2025,
  title   = {{LitBench}: A Benchmark and Dataset for Reliable Evaluation of Creative Writing},
  author  = {Fein, Daniel and Russo, Sebastian and Xiang, Violet and Jolly, Kabir and Rafailov, Rafael and Haber, Nick},
  booktitle = {Proceedings of the 19th Conference of the European Chapter of the Association for Computational Linguistics (Volume 1: Long Papers)},
  pages   = {7740--7755},
  year    = {2026}
}

@inproceedings{hanna2022,
  title     = {Of Human Criteria and Automatic Metrics: A Benchmark of the Evaluation of Story Generation},
  author    = {Chhun, Cyril and Colombo, Pierre and Suchanek, Fabian M. and Clavel, Chlo\'{e}},
  booktitle = {Proceedings of the 29th International Conference on Computational Linguistics},
  year      = {2022},
  pages     = {5794--5836},
  publisher = {International Committee on Computational Linguistics}
}

@article{goodstorysurvey2024,
  title   = {What Makes a Good Story and How Can We Measure It? A Comprehensive Survey of Story Evaluation},
  author  = {Yang, Dingyi and Jin, Qin},
  journal = {arXiv preprint arXiv:2408.14622},
  year    = {2024},
  url     = {https://arxiv.org/abs/2408.14622}
}

@inproceedings{storyer2022,
  title   = {{StoryER}: Automatic Story Evaluation via Ranking, Rating and Reasoning},
  author  = {Chen, Hong and Vo, Duc Minh and Takamura, Hiroya and Miyao, Yusuke and Nakayama, Hideki},
  booktitle = {Proceedings of the 2022 Conference on Empirical Methods in Natural Language Processing},
  pages   = {1739--1753},
  year    = {2022}
}

@article{measuringaislop2025,
  title   = {Measuring {AI} ``Slop'' in Text},
  author  = {Shaib, Chantal and Chakrabarty, Tuhin and Garcia-Olano, Diego and Wallace, Byron C.},
  journal = {arXiv preprint arXiv:2509.19163},
  year    = {2025},
  url     = {https://arxiv.org/abs/2509.19163}
}

@inproceedings{geval2023,
  title   = {{G-Eval}: {NLG} Evaluation using {GPT-4} with Better Human Alignment},
  author  = {Liu, Yang and Iter, Dan and Xu, Yichong and Wang, Shuohang and Xu, Ruochen and Zhu, Chenguang},
  booktitle = {Proceedings of the 2023 Conference on Empirical Methods in Natural Language Processing},
  pages   = {2511--2522},
  year    = {2023}
}

@inproceedings{prometheus2024,
  title   = {Prometheus 2: An Open Source Language Model Specialized in Evaluating Other Language Models},
  author  = {Kim, Seungone and Suk, Juyoung and Longpre, Shayne and Lin, Bill Yuchen and Shin, Jamin and Welleck, Sean and Neubig, Graham and Lee, Moontae and Lee, Kyungjae and Seo, Minjoon},
  booktitle = {Proceedings of the 2024 Conference on Empirical Methods in Natural Language Processing},
  pages   = {4334--4353},
  year    = {2024}
}

@article{nofreelabels2025,
  title   = {No Free Labels: Limitations of {LLM}-as-a-Judge Without Human Grounding},
  author  = {Krumdick, Michael and Lovering, Charles and Reddy, Varshini and Ebner, Seth and Tanner, Chris},
  journal = {arXiv preprint arXiv:2503.05061},
  year    = {2025},
  url     = {https://arxiv.org/abs/2503.05061}
}

@article{judgmentnoise2025,
  title   = {When Judgment Becomes Noise: How Design Failures in {LLM} Judge Benchmarks Silently Undermine Validity},
  author  = {Feuer, Benjamin and Tseng, Chiung-Yi and Lathe, Astitwa Sarthak and Elachqar, Oussama and Dickerson, John P.},
  journal = {arXiv preprint arXiv:2509.20293},
  year    = {2025},
  url     = {https://arxiv.org/abs/2509.20293}
}

@inproceedings{antislop2025,
  title   = {Antislop: A Comprehensive Framework for Identifying and Eliminating Repetitive Patterns in Language Models},
  author  = {Paech, Sam and Roush, Allen and Goldfeder, Judah and Shwartz-Ziv, Ravid},
  booktitle = {International Conference on Learning Representations},
  year    = {2026}
}

@inproceedings{translationesedialects2011,
  title     = {Translationese and Its Dialects},
  author    = {Koppel, Moshe and Ordan, Noam},
  booktitle = {Proceedings of the 49th Annual Meeting of the Association for Computational Linguistics: Human Language Technologies},
  year      = {2011},
  pages     = {1318--1326},
  publisher = {Association for Computational Linguistics}
}

@article{translationeseunsup2015,
  title   = {Unsupervised Identification of Translationese},
  author  = {Rabinovich, Ella and Wintner, Shuly},
  journal = {Transactions of the Association for Computational Linguistics},
  volume  = {3},
  year    = {2015},
  pages   = {419--432},
  publisher = {MIT Press}
}

@inproceedings{chatbotarena2024,
  title     = {{Chatbot Arena}: An Open Platform for Evaluating {LLMs} by Human Preference},
  author    = {Chiang, Wei-Lin and Zheng, Lianmin and Sheng, Ying and Angelopoulos, Anastasios Nikolas and Li, Tianle and Li, Dacheng and Zhu, Banghua and Zhang, Hao and Jordan, Michael I. and Gonzalez, Joseph E. and Stoica, Ion},
  booktitle = {Forty-first International Conference on Machine Learning (ICML)},
  year      = {2024},
  url       = {https://arxiv.org/abs/2403.04132}
}

@inproceedings{indicnlg2022,
  title   = {{IndicNLG} Benchmark: Multilingual Datasets for Diverse {NLG} Tasks in {Indic} Languages},
  author  = {Kumar, Aman and Shrotriya, Himani and Sahu, Prachi and Mishra, Amogh and Dabre, Raj and Puduppully, Ratish and Kunchukuttan, Anoop and Khapra, Mitesh M. and Kumar, Pratyush},
  booktitle = {Proceedings of the 2022 Conference on Empirical Methods in Natural Language Processing},
  pages   = {5363--5394},
  year    = {2022}
}

@inproceedings{milu2024,
  title     = {{MILU}: A Multi-task {Indic} Language Understanding Benchmark},
  author    = {Verma, Sshubam and Khan, Mohammed Safi Ur Rahman and Kumar, Vishwajeet and Murthy, Rudra and Sen, Jaydeep},
  booktitle = {Proceedings of the 2025 Conference of the Nations of the Americas Chapter of the Association for Computational Linguistics: Human Language Technologies (Volume 1: Long Papers)},
  year      = {2025},
  pages     = {10076--10132},
  publisher = {Association for Computational Linguistics},
  doi       = {10.18653/v1/2025.naacl-long.507},
  url       = {https://aclanthology.org/2025.naacl-long.507/}
}

@article{airavata2024,
  title   = {Airavata: Introducing {Hindi} Instruction-tuned {LLM}},
  author  = {Gala, Jay and Jayakumar, Thanmay and Husain, Jaavid Aktar and Kumar M, Aswanth and Khan, Mohammed Safi Ur Rahman and Kanojia, Diptesh and Puduppully, Ratish and Khapra, Mitesh M. and Dabre, Raj and Murthy, Rudra and Kunchukuttan, Anoop},
  journal = {arXiv preprint arXiv:2401.15006},
  year    = {2024},
  url     = {https://arxiv.org/abs/2401.15006}
}

@inproceedings{coedit2023,
  title   = {{CoEdIT}: Text Editing by Task-Specific Instruction Tuning},
  author  = {Raheja, Vipul and Kumar, Dhruv and Koo, Ryan and Kang, Dongyeop},
  booktitle = {Findings of the Association for Computational Linguistics: EMNLP 2023},
  pages   = {5274--5291},
  year    = {2023}
}

@article{selfrefine2023,
  title   = {{Self-Refine}: Iterative Refinement with Self-Feedback},
  author  = {Madaan, Aman and Tandon, Niket and Gupta, Prakhar and Hallinan, Skyler and Gao, Luyu and Wiegreffe, Sarah and Alon, Uri and Dziri, Nouha and Prabhumoye, Shrimai and Yang, Yiming and Gupta, Shashank and Majumder, Bodhisattwa Prasad and Hermann, Katherine and Welleck, Sean and Yazdanbakhsh, Amir and Clark, Peter},
  journal = {Advances in Neural Information Processing Systems},
  volume  = {36},
  pages   = {46534--46594},
  year    = {2023}
}

@article{summac2022,
  title   = {{SummaC}: Re-Visiting {NLI}-based Models for Inconsistency Detection in Summarization},
  author  = {Laban, Philippe and Schnabel, Tobias and Bennett, Paul N. and Hearst, Marti A.},
  journal = {Transactions of the Association for Computational Linguistics},
  volume  = {10},
  year    = {2022},
  pages   = {163--177},
  publisher = {MIT Press}
}

@inproceedings{translationesenmt2020,
  title     = {Translationese as a Language in ``Multilingual'' {NMT}},
  author    = {Riley, Parker and Caswell, Isaac and Freitag, Markus and Grangier, David},
  booktitle = {Proceedings of the 58th Annual Meeting of the Association for Computational Linguistics (ACL)},
  year      = {2020},
  pages     = {7737--7746}
}

@inproceedings{storygensurvey2025,
  title     = {A Survey on {LLMs} for Story Generation},
  author    = {Teleki, Maria and Bengali, Vedangi and Dong, Xiangjue and Janjur, Sai Tejas and Liu, Haoran and Liu, Tian and Wang, Cong and Liu, Ting and Zhang, Yin and Shipman, Frank and Caverlee, James},
  booktitle = {Findings of the Association for Computational Linguistics: EMNLP 2025},
  year      = {2025},
  pages     = {13954--13966},
  publisher = {Association for Computational Linguistics},
  doi       = {10.18653/v1/2025.findings-emnlp.750},
  url       = {https://aclanthology.org/2025.findings-emnlp.750/}
}

@article{dome2024,
  title   = {Generating Long-form Story Using Dynamic Hierarchical Outlining with Memory-Enhancement},
  author  = {Wang, Qianyue and Hu, Jinwu and Li, Zhengping and Wang, Yufeng and Li, Daiyuan and Hu, Yu and Tan, Mingkui},
  journal = {arXiv preprint arXiv:2412.13575},
  year    = {2024},
  url     = {https://arxiv.org/abs/2412.13575}
}

@article{10.1093/biomet/39.3-4.324,
  author  = {Bradley, Ralph Allan and Terry, Milton E.},
  title   = {Rank Analysis of Incomplete Block Designs: The Method of Paired Comparisons},
  journal = {Biometrika},
  volume  = {39},
  number  = {3--4},
  pages   = {324--345},
  year    = {1952},
  month   = dec,
  issn    = {0006-3444},
  doi     = {10.1093/biomet/39.3-4.324},
  url     = {https://doi.org/10.1093/biomet/39.3-4.324}
}

@inproceedings{wang-etal-2024-large-language-models-fair,
  title     = {Large Language Models are not Fair Evaluators},
  author    = {Wang, Peiyi and Li, Lei and Chen, Liang and Cai, Zefan and Zhu, Dawei and Lin, Binghuai and Cao, Yunbo and Kong, Lingpeng and Liu, Qi and Liu, Tianyu and Sui, Zhifang},
  editor    = {Ku, Lun-Wei and Martins, Andre and Srikumar, Vivek},
  booktitle = {Proceedings of the 62nd Annual Meeting of the Association for Computational Linguistics (Volume 1: Long Papers)},
  month     = aug,
  year      = {2024},
  address   = {Bangkok, Thailand},
  publisher = {Association for Computational Linguistics},
  url       = {https://aclanthology.org/2024.acl-long.511/},
  doi       = {10.18653/v1/2024.acl-long.511},
  pages     = {9440--9450}
}

@article{kumar2019bhaav,
  title   = {{BHAAV}: A Text Corpus for Emotion Analysis from {Hindi} Stories},
  author  = {Kumar, Yaman and Mahata, Debanjan and Aggarwal, Sagar and Chugh, Anmol and Maheshwari, Rajat and Shah, Rajiv Ratn},
  journal = {arXiv preprint arXiv:1910.04073},
  year    = {2019}
}

@inproceedings{dhanwal2020discourse,
  title     = {An Annotated Dataset of Discourse Modes in {Hindi} Stories},
  author    = {Dhanwal, Swapnil and Dutta, Hritwik and Nankani, Hitesh and Shrivastava, Nilay and Kumar, Yaman and Li, Junyi Jessy and Mahata, Debanjan and Gosangi, Rakesh and Zhang, Haimin and Shah, Rajiv Ratn and Stent, Amanda},
  booktitle = {Proceedings of the Twelfth Language Resources and Evaluation Conference},
  year      = {2020},
  pages     = {1191--1196},
  url       = {https://aclanthology.org/2020.lrec-1.149/}
}

@misc{openai2026gpt55,
  author = {{OpenAI}},
  title  = {{GPT-5.5} System Card},
  year   = {2026},
  month  = apr,
  url    = {https://openai.com/index/gpt-5-5-system-card/}
}

@misc{anthropic2026opus46,
  author = {{Anthropic}},
  title  = {{Claude Opus 4.6} System Card},
  year   = {2026},
  month  = feb,
  url    = {https://anthropic.com/claude-opus-4-6-system-card}
}

@misc{google2026gemini31pro,
  author = {{Google DeepMind}},
  title  = {{Gemini 3.1 Pro} Model Card},
  year   = {2026},
  month  = feb,
  url    = {https://deepmind.google/models/model-cards/gemini-3-1-pro/}
}

@article{deepseek2026v4,
  author  = {{DeepSeek-AI}},
  title   = {{DeepSeek-V4}: Towards Highly Efficient Million-Token Context Intelligence},
  journal = {arXiv preprint arXiv:2606.19348},
  year    = {2026},
  month   = apr,
  url     = {https://arxiv.org/abs/2606.19348}
}

@article{google2026gemma4,
  author  = {{Gemma Team}},
  title   = {Gemma 4 Technical Report},
  journal = {arXiv preprint arXiv:2607.02770},
  year    = {2026},
  month   = jul,
  url     = {https://arxiv.org/abs/2607.02770}
}

@misc{sarvam2026sovereign,
  author = {{Sarvam Foundation Models Team}},
  title  = {Introducing {Sarvam's} Sovereign Models},
  year   = {2026},
  url    = {https://huggingface.co/sarvamai/sarvam-105b}
}

@misc{qwen2026qwen35,
  author = {{Qwen Team}},
  title  = {{Qwen3.5}: Towards Native Multimodal Agents},
  year   = {2026},
  month  = feb,
  url    = {https://huggingface.co/Qwen/Qwen3.5-27B}
}

@misc{xai2026grok43,
  author = {{xAI}},
  title  = {Grok on {Amazon Bedrock}},
  year   = {2026},
  month  = jun,
  url    = {https://x.ai/news/grok-amazon-bedrock}
}

@article{moonshot2026kimik3,
  author  = {{Kimi Team}},
  title   = {{Kimi K3}: Open Frontier Intelligence},
  journal = {arXiv preprint arXiv:2607.24653},
  year    = {2026},
  url     = {https://arxiv.org/abs/2607.24653}
}

@misc{google2026gemini31flashlite,
  author = {{Google DeepMind}},
  title  = {{Gemini 3.1 Flash-Lite} Model Card},
  year   = {2026},
  month  = mar,
  url    = {https://deepmind.google/models/model-cards/gemini-3-1-flash-lite/}
}

@misc{openai2026gpt54mini,
  author = {{OpenAI}},
  title  = {Introducing {GPT-5.4} Mini and Nano},
  year   = {2026},
  month  = mar,
  url    = {https://openai.com/index/introducing-gpt-5-4-mini-and-nano/}
}

@misc{anthropic2025sonnet45,
  author = {{Anthropic}},
  title  = {{Claude Sonnet 4.5} System Card},
  year   = {2025},
  month  = sep,
  url    = {https://anthropic.com/claude-sonnet-4-5-system-card}
}

@misc{google2025gemini3flash,
  author = {{Google DeepMind}},
  title  = {{Gemini 3 Flash} Model Card},
  year   = {2025},
  month  = dec,
  url    = {https://deepmind.google/models/model-cards/gemini-3-flash/}
}

@article{moonshot2025kimik2,
  author  = {{Kimi Team}},
  title   = {{Kimi K2}: Open Agentic Intelligence},
  journal = {arXiv preprint arXiv:2507.20534},
  year    = {2025},
  url     = {https://arxiv.org/abs/2507.20534}
}

@article{glm5team2026,
  author  = {{GLM-5-Team}},
  title   = {{GLM-5}: from Vibe Coding to Agentic Engineering},
  journal = {arXiv preprint arXiv:2602.15763},
  year    = {2026},
  url     = {https://arxiv.org/abs/2602.15763}
}

@misc{minimax2026m27,
  author = {{MiniMax}},
  title  = {{MiniMax M2.7}: Early Echoes of Self-Evolution},
  year   = {2026},
  month  = mar,
  url    = {https://www.minimax.io/news/minimax-m27-en}
}

@misc{mistral2026medium35,
  author = {{Mistral AI}},
  title  = {{Mistral Medium 3.5}},
  year   = {2026},
  month  = apr,
  url    = {https://docs.mistral.ai/models/mistral-medium-3-5-26-04}
}

@inproceedings{pham-etal-2026-analystbench,
    title = "{A}nalyst{B}ench: Benchmarking professional long-form report generation with web-mined multimodal tasks",
    author = "Pham, Chau Minh  and
      Wang, Zichao  and
      Mathur, Puneet  and
      Siu, Alexa  and
      Jain, Akriti  and
      Garimella, Aparna  and
      Sai, Ananya B.  and
      Lipka, Nedim  and
      Iyyer, Mohit  and
      Manjunatha, Varun",
    editor = "Liakata, Maria  and
      Moreira, Viviane P.  and
      Zhang, Jiajun  and
      Jurgens, David",
    booktitle = "Findings of the {A}ssociation for {C}omputational {L}inguistics: {ACL} 2026",
    month = jul,
    year = "2026",
    address = "San Diego, California, United States",
    publisher = "Association for Computational Linguistics",
    url = "https://aclanthology.org/2026.findings-acl.1197/",
    doi = "10.18653/v1/2026.findings-acl.1197",
    pages = "23894--23926",
    ISBN = "979-8-89176-395-1"
}

@inproceedings{kazi-etal-2026-adaptation,
    title = "Adaptation, Not Translation: An Architecture and Evaluation Framework for Long-Form Cross-Cultural Story Localization",
    author = "Kazi, Taaha  and
      Kanade, Aditya Sanjiv  and
      Yang, Ivory  and
      Saifullah, Mohammad  and
      Sharma, Vasu  and
      Pant, Bhavesh  and
      Bhatia, Kunal  and
      Prakash, Abhinav  and
      Rahman, Abdur  and
      Agarwal, Varun  and
      Jain, Sarjita  and
      Pavate, Viveca  and
      Maithul, Nishita",
    editor = "Atanasova, Pepa  and
      Haffari, Gholamreza  and
      Wang, Minghan",
    booktitle = "Proceedings of the 2026 Conference on Empirical Methods in Natural Language Processing: Industry Track",
    month = oct,
    year = "2026",
    address = "Budapest, Hungary",
    publisher = "Association for Computational Linguistics",
    url = "https://aclanthology.org/2026.emnlp-industry.122/",
    doi = "10.18653/2026.emnlp-industry.122",
    pages = "2093--2114",
    ISBN = "979-8-89176-460-6"
}

@inproceedings{rahman-etal-2026-calques,
    title = "Why Calques Look Rare in {H}indi Model Output: A Human-Anchored Measurement Study",
    author = "Rahman, Abdur  and
      Sharma, Vasu  and
      Saifullah, Mohammad",
    editor = "Ranaldi, Leonardo  and
      Pucci, Giulia  and
      Ranaldi, Federico  and
      Valentino, Marco  and
      Birch, Alexandra  and
      Zanzotto, Fabio Massimo",
    booktitle = "The Proceedings of the Workshop on Open Reasoning Across Cultures Languages ({ORACLE})",
    month = oct,
    year = "2026",
    address = "Budapest, Hungary",
    publisher = "Association for Computational Linguistics",
    url = "https://aclanthology.org/2026.oracle-main.21/",
    doi = "10.18653/2026.oracle-main.21",
    pages = "316--328",
    ISBN = "979-8-89176-483-5"
}

@inproceedings{mann2026narrativeworldbench,
title={{NarrativeWorldBench}: A Frontier-Saturated Benchmark and a Latent World Model for Long-Horizon Co-Creative Audio Drama},
author={Logan Mann and Abdur Rahman and Mohammad Saifullah and Taaha Kazi and Vasu Sharma},
booktitle={Culture x AI: Evaluating AI as a Cultural Technology (ICML 2026)},
year={2026},
url={https://openreview.net/forum?id=Zxop1N6fyh}
}

@misc{kazi2026pauseeditablestrategyartifacts,
      title={{PAUSE}: Editable Strategy Artifacts for Long-Form Cultural Story Adaptation}, 
      author={Taaha Kazi and Vasu Sharma and Mohammad Saifullah and Abdur Rahman},
      year={2026},
      eprint={2608.28633},
      archivePrefix={arXiv},
      primaryClass={cs.CL},
      url={https://arxiv.org/abs/2608.28633}
}

@misc{saifullah2026narrativeworldmodelnarratologygrounded,
      title={Narrative World Model: Narratology-Grounded Writer Memory for Long-Form Fiction}, 
      author={Mohammad Saifullah and Thomas Kornmaier and Taaha Kazi and Vasu Sharma and Aditya Sanjiv Kanade and Aanand Kumar Yadav},
      year={2026},
      eprint={2607.05577},
      archivePrefix={arXiv},
      primaryClass={cs.AI},
      url={https://arxiv.org/abs/2607.05577}
}

@misc{hindiwikisource,
  author = {{Hindi Wikisource contributors}},
  title  = {{Hindi Wikisource}},
  url    = {https://hi.wikisource.org/},
}

\appendix
\let\FloatBarrier\relax

\section{Per-form headline leaderboards}\label{app:perform}
Table~\ref{tab:perform} gives the form-specific headline tables. Bradley--Terry is fit
separately within each form; the table shows raw centered point estimates, and the
release provides their 95\% source-bootstrap intervals. The standardized macro interval
is in Table~\ref{tab:leaderboard}. Win is the faithfulness-gated empirical win rate,
Fail the output-level gate-failure rate, and J1--J3 the form-conditioned 1--5 diagnostic
means. Table~\ref{tab:leaderboard} divides each form's centered coefficient vector by its population SD
before averaging the three standardized values; the raw form-specific coefficients
remain here. The middle of the table reorders across forms
(for example, Gemini 3.1 Pro is third in short stories but fourth in essays, and
DeepSeek V4-Pro is fourth in short stories but third in audio fiction and essays).

\begin{table}[htbp]\centering\small\setlength{\tabcolsep}{2pt}
\textbf{A. Audio fiction}\\[2pt]
\begin{tabular}{@{}rlrrrrrr@{}}
\toprule
Rk & Model & BT & Win & Fail & J1 & J2 & J3 \\
\midrule
1 & GPT-5.5 & 2.02 & 0.891 & 13.4\% & 3.95 & 4.10 & 3.87 \\
2 & Claude Opus 4-6 & 1.13 & 0.752 & 21.2\% & 3.76 & 4.12 & 3.83 \\
3 & DeepSeek V4-Pro & 0.28 & 0.578 & 12.9\% & 3.60 & 3.99 & 3.66 \\
4 & Gemma 4 26B & -0.46 & 0.399 & 42.8\% & 3.64 & 3.97 & 3.66 \\
5 & Gemini 3.1 Pro & -0.57 & 0.373 & 57.5\% & 3.68 & 4.02 & 3.73 \\
6 & Sarvam 105B & -0.69 & 0.350 & 14.8\% & 3.27 & 3.95 & 3.52 \\
7 & Grok 4.3 & -0.71 & 0.341 & 45.8\% & 3.70 & 4.06 & 3.70 \\
8 & Qwen3.5 27B & -1.01 & 0.277 & 27.6\% & 3.01 & 3.89 & 3.54 \\
\bottomrule
\end{tabular}\\[6pt]
\textbf{B. Short stories}\\[2pt]
\begin{tabular}{@{}rlrrrrrr@{}}
\toprule
Rk & Model & BT & Win & Fail & J1 & J2 & J3 \\
\midrule
1 & GPT-5.5 & 2.03 & 0.877 & 3.0\% & 3.93 & 3.46 & 3.41 \\
2 & Claude Opus 4-6 & 1.59 & 0.814 & 8.2\% & 3.82 & 3.50 & 3.41 \\
3 & Gemini 3.1 Pro & 0.36 & 0.587 & 14.6\% & 3.72 & 3.35 & 3.29 \\
4 & DeepSeek V4-Pro & -0.16 & 0.476 & 3.8\% & 3.40 & 3.33 & 3.24 \\
5 & Grok 4.3 & -0.26 & 0.456 & 12.0\% & 3.41 & 3.34 & 3.26 \\
6 & Gemma 4 26B & -0.72 & 0.356 & 15.4\% & 3.39 & 3.26 & 3.19 \\
7 & Qwen3.5 27B & -1.38 & 0.223 & 7.2\% & 3.11 & 3.27 & 3.19 \\
8 & Sarvam 105B & -1.46 & 0.209 & 8.8\% & 3.13 & 3.24 & 3.16 \\
\bottomrule
\end{tabular}\\[6pt]
\textbf{C. Literary essays}\\[2pt]
\begin{tabular}{@{}rlrrrrrr@{}}
\toprule
Rk & Model & BT & Win & Fail & J1 & J2 & J3 \\
\midrule
1 & GPT-5.5 & 1.80 & 0.852 & 6.6\% & 3.97 & 3.11 & 3.49 \\
2 & Claude Opus 4-6 & 1.48 & 0.803 & 12.3\% & 3.89 & 3.04 & 3.39 \\
3 & DeepSeek V4-Pro & 0.32 & 0.578 & 10.0\% & 3.64 & 2.96 & 3.31 \\
4 & Gemini 3.1 Pro & 0.18 & 0.545 & 29.4\% & 3.84 & 3.00 & 3.32 \\
5 & Grok 4.3 & -0.33 & 0.432 & 24.3\% & 3.73 & 2.98 & 3.33 \\
6 & Gemma 4 26B & -0.86 & 0.318 & 25.1\% & 3.63 & 2.95 & 3.28 \\
7 & Sarvam 105B & -1.22 & 0.245 & 9.1\% & 3.42 & 2.93 & 3.24 \\
8 & Qwen3.5 27B & -1.37 & 0.216 & 20.0\% & 3.40 & 2.94 & 3.28 \\
\bottomrule
\end{tabular}
\caption{Form-specific headline leaderboards. See Table~\ref{tab:leaderboard} for the
equal-form standardized macro derived from these raw form-specific fits.}
\label{tab:perform}
\end{table}

\FloatBarrier

\section{Leaderboard robustness}\label{app:robustness}

\textbf{Position bias.} Each pair is judged in both presentation orders and collapsed to
a single source-level outcome; this swap-collapse neutralizes presentation-order bias by
construction, and any two-order disagreement is collapsed to a tie.

\begin{figure*}[htbp]\centering
\includegraphics[width=0.9\textwidth]{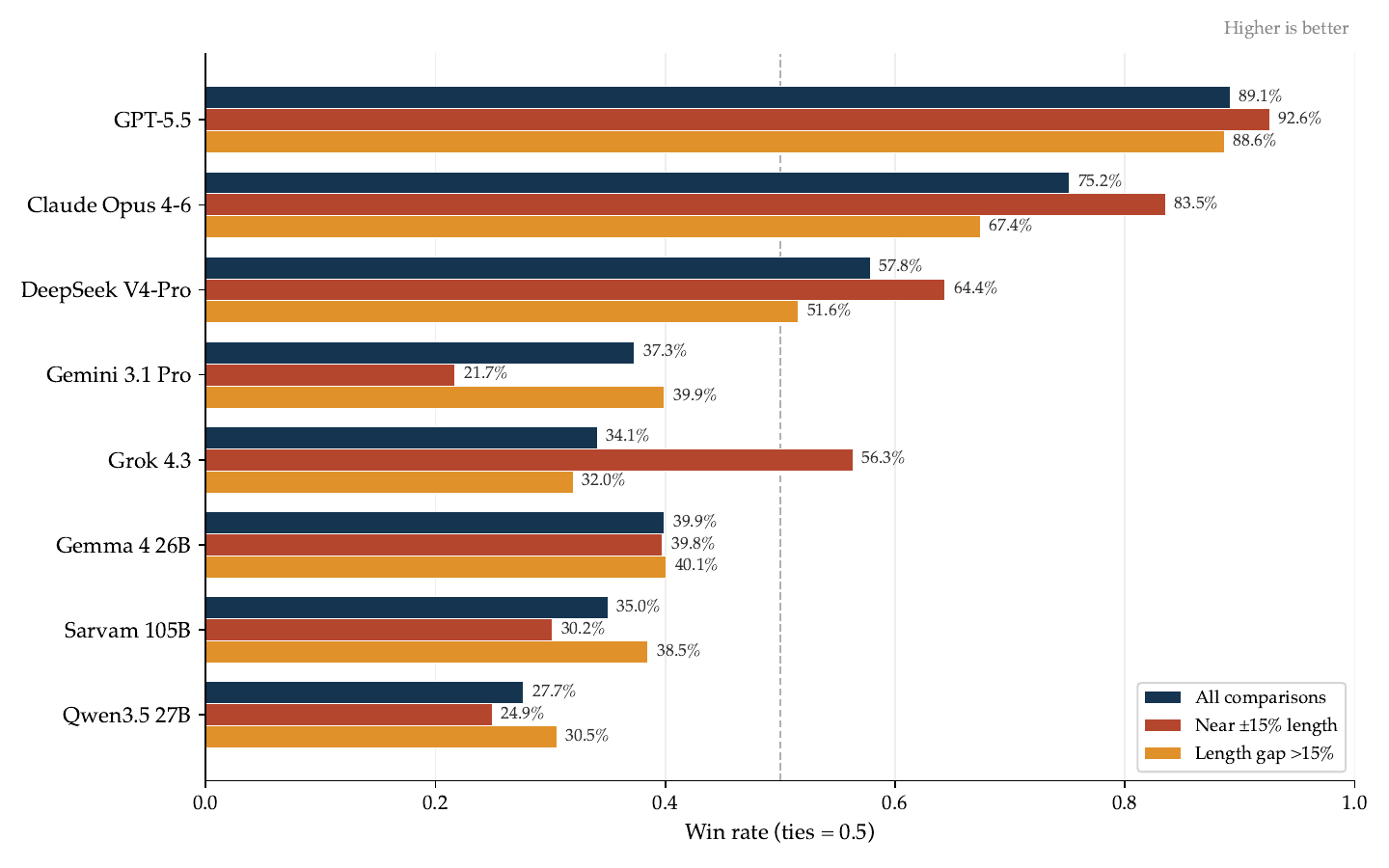}
\caption{Audio-fiction length-control sensitivity: empirical win rates over all resolved
comparisons, near-length comparisons, and comparisons with a larger output-length gap. The
top two models remain unchanged under near-length matching, while some lower ranks move.
Together with Table~\ref{tab:lengthcontrol}, this shows that the headline leaders are stable
across all three forms and that the complete written-form rankings are preserved.}
\label{fig:length}
\end{figure*}

\textbf{Length control.} A faithful refinement can expand its draft. Near-length matching
preserves GPT-5.5 and Claude Opus 4-6 as the top two models in every form and preserves the
complete short-story and essay rankings (Table~\ref{tab:lengthcontrol}). Audio fiction is
more sensitive below the top three, so Figure~\ref{fig:length} reports its unrestricted,
near-length, and large-gap results separately.

\raggedbottom
\begin{table*}[htbp]\centering\small
\begin{tabular}{@{}lccc@{}}
\toprule
Model & Audio WR ($n$) & Story WR ($n$) & Essay WR ($n$) \\
\midrule
GPT-5.5 & \textbf{92.6\% (1{,}056)} & \textbf{87.8\% (1{,}824)} & \textbf{84.4\% (1{,}664)} \\
Claude Opus 4-6 & \textbf{83.5\% (3{,}751)} & \textbf{85.1\% (2{,}407)} & \textbf{83.1\% (2{,}060)} \\
DeepSeek V4-Pro & 64.4\% (3{,}926) & 51.0\% (2{,}700) & 59.9\% (2{,}072) \\
Gemini 3.1 Pro & 21.7\% (1{,}011) & 60.6\% (2{,}034) & 56.1\% (1{,}597) \\
Grok 4.3 & 56.3\% (633) & 45.7\% (1{,}431) & 45.8\% (1{,}494) \\
Gemma 4 26B & 39.8\% (3{,}614) & 37.8\% (2{,}624) & 31.8\% (2{,}059) \\
Sarvam 105B & 30.2\% (3{,}345) & 23.1\% (2{,}592) & 23.8\% (2{,}040) \\
Qwen3.5 27B & 24.9\% (3{,}956) & 23.6\% (2{,}646) & 21.1\% (2{,}026) \\
\bottomrule
\end{tabular}
\caption{Near-length empirical win rates on resolved comparisons with an
output-word-count ratio in $[0.85, 1.15]$. Ties
count as 0.5; both-fail and invalid outcomes are excluded. $n$ is the number of
model-comparison exposures. This is a descriptive sensitivity analysis, not a
length-adjusted leaderboard.}
\label{tab:lengthcontrol}
\end{table*}

\FloatBarrier

\FloatBarrier

\textbf{Which pairings length matching keeps.} Near-length matching changes which
opponents a model meets. On audio fiction this explains Gemini's large move fully and
Grok's in part (Table~\ref{tab:lengthcontrol} and Table~\ref{tab:lengthopp}). Gemini 3.1 Pro and GPT-5.5
expand audio drafts the most (1.661 and 1.598 times the draft;
Appendix~\ref{app:repro}), so 691 of Gemini's 1{,}011 near-length comparisons are
against GPT-5.5, a pairing Gemini loses at any length (0.054 near-length, 0.055
overall). Against each other opponent it meets more than once, Gemini's near-length rate
is at or above its overall rate. Holding its overall per-opponent rates fixed and
reweighting them to the near-length mix gives 0.167, below the observed 0.217, so the
change of opponents accounts for the whole drop from 0.373. Grok 4.3, the shortest audio
refiner (0.775 times the draft), has no near-length comparison against GPT-5.5, and 339
of its 633 are against Sarvam 105B. The same reweighting gives 0.433, so the mix
accounts for about two-fifths of its rise from 0.341 to 0.563; the rest comes from
higher rates within pairings, mainly against Sarvam 105B and Qwen3.5 27B. In the written
forms both models stay between 0.84 and 1.12 times the draft, and their near-length
rates stay within 0.03 of their overall rates under the same judge, which points to the
models' audio length behavior. The decomposition is descriptive.

\begin{table}[htbp]\centering\small\setlength{\tabcolsep}{3pt}
\begin{tabular}{@{}lrrrr@{}}
\toprule
& \multicolumn{2}{c}{Gemini 3.1 Pro} & \multicolumn{2}{c}{Grok 4.3} \\
\cmidrule(lr){2-3}\cmidrule(lr){4-5}
Opponent & Near ($n$) & All & Near ($n$) & All \\
\midrule
GPT-5.5 & 0.054 (691) & 0.055 & (0) & 0.071 \\
Claude Opus 4-6 & 0.192 (73) & 0.138 & 0.103 (68) & 0.123 \\
DeepSeek V4-Pro & 0.500 (71) & 0.407 & 0.338 (145) & 0.252 \\
Gemini 3.1 Pro & & & 0.000 (1) & 0.440 \\
Grok 4.3 & 1.000 (1) & 0.560 & & \\
Gemma 4 26B & 0.734 (77) & 0.552 & 0.380 (25) & 0.387 \\
Sarvam 105B & 0.893 (14) & 0.467 & 0.724 (339) & 0.548 \\
Qwen3.5 27B & 0.750 (84) & 0.506 & 0.827 (55) & 0.606 \\
\midrule
All & 0.217 (1{,}011) & 0.373 & 0.563 (633) & 0.341 \\
\bottomrule
\end{tabular}
\caption{Audio-fiction win rates of Gemini 3.1 Pro and Grok 4.3 by opponent, over
near-length comparisons (word ratio in $[0.85, 1.15]$) and over all resolved
comparisons. Ties count one half.}
\label{tab:lengthopp}
\end{table}

\FloatBarrier

\textbf{Both-pass prose-only diagnostic.} Restricting to comparisons where both rewrites
passed the gate gives the prose-quality order among faithful outputs
(Table~\ref{tab:bothpass}). This is a diagnostic that excludes gate-failed
outputs; the ungated fit in Appendix~\ref{app:ablations} prose-judges them as well. GPT-5.5 and Claude Opus 4-6 stay first and
second in every form, and Gemini 3.1 Pro rises to third overall, consistent with strong
prose when it stays faithful.

\begin{table}[htbp]\centering\small
\begin{tabular}{@{}rlrrrr@{}}
\toprule
Rk & Model & Macro & Audio & Story & Essay \\
\midrule
1 & GPT-5.5 & 0.910 & 0.972 & 0.888 & 0.871 \\
2 & Claude Opus 4-6 & 0.854 & 0.837 & 0.866 & 0.860 \\
3 & Gemini 3.1 Pro & 0.665 & 0.663 & 0.648 & 0.684 \\
4 & DeepSeek V4-Pro & 0.483 & 0.466 & 0.438 & 0.546 \\
5 & Grok 4.3 & 0.439 & 0.388 & 0.466 & 0.463 \\
6 & Gemma 4 26B & 0.380 & 0.471 & 0.366 & 0.304 \\
7 & Sarvam 105B & 0.125 & 0.113 & 0.152 & 0.111 \\
8 & Qwen3.5 27B & 0.121 & 0.073 & 0.161 & 0.128 \\
\bottomrule
\end{tabular}
\caption{Both-pass prose-only empirical win rate (gate-passing comparisons only), equal-
form macro and per form. A selected subset that explains the contribution of prose
quality among faithful outputs; not an ungated leaderboard.}
\label{tab:bothpass}
\end{table}
\flushbottom

\FloatBarrier

\section{Ablations}\label{app:ablations}
Both ablations use the published judging exactly: GPT-5.5 under the frozen headline
prompt, both presentation orders collapsed, ties as half a win, and the per-form fit,
center, scale, and equal-weight average of the headline, with 1{,}000 source-item
bootstrap replicates. They add 38{,}614 new pairs (77{,}228 judgments over the two orders)
to the published run.

\textbf{Gate ablation.} The published run prose-judges only pairs in which both
rewrites pass the gate. We prose-judged the remaining 22{,}214 pairs (16{,}867 audio
fiction, 2{,}395 short story, 2{,}952 essay), so the judge covers every one of the
57{,}400 same-source pairs, and then varied the gate by its violation-tag count, its
only severity signal: an output fails at threshold $t$ if it carries at least $t$
violation tags. As a control, $t=1$ reproduces the published leaderboard and every
dropped and gate-decided pair count exactly. Table~\ref{tab:ablation} shows that the
order holds at $t=2$, and that at $t=3$ and with no gate only Gemini 3.1 Pro and
DeepSeek V4-Pro swap third and fourth (Spearman 0.976; ungated, adjacent models
still separate in every bootstrap replicate). The ungated condition is an ablation of our own design, reported
for diagnosis only: the prose judge never sees the draft, so without the gate nothing
checks fidelity, and the model that rises is the one with the highest gate failure rate.

\textbf{Absolute anchor: the draft as a candidate.} We also judged each draft against
each of its eight refinements (16{,}400 pairs). The draft has no gate verdict of its
own, since the gate compares a refinement against its draft, so it can enter only the
ungated fit, where it is a ninth candidate (Table~\ref{tab:ablation}, last column).
Table~\ref{tab:drafts} gives each refiner's direct win rate against its own draft. Six
refiners are preferred to the draft in all three forms, at 0.748 or above. Sarvam 105B
wins narrowly in each form but cannot be separated from the draft in the nine-candidate fit
($P=0.834$, overlapping intervals), and Qwen3.5 27B falls below the draft on audio
fiction and in that fit. The draft is therefore a genuine competitor.

\begin{table}[htbp]\centering\footnotesize\setlength{\tabcolsep}{2.5pt}
\resizebox{\columnwidth}{!}{%
\begin{tabular}{@{}lrrrrr@{}}
\toprule
Model & $t{\geq}1$ & $t{\geq}2$ & $t{\geq}3$ & None & +Draft \\
\midrule
GPT-5.5 & 1.783 & 1.675 & 1.602 & 1.545 & 1.673 \\
Claude Opus 4-6 & 1.268 & 1.278 & 1.283 & 1.271 & 1.397 \\
DeepSeek V4-Pro & 0.146 & 0.123 & $-0.007$ & $-0.041$ & 0.073 \\
Gemini 3.1 Pro & $-0.033$ & 0.108 & 0.285 & 0.399 & 0.496 \\
Grok 4.3 & $-0.409$ & $-0.348$ & $-0.212$ & $-0.136$ & $-0.023$ \\
Gemma 4 26B & $-0.613$ & $-0.537$ & $-0.461$ & $-0.427$ & $-0.294$ \\
Sarvam 105B & $-1.005$ & $-1.112$ & $-1.221$ & $-1.274$ & $-1.085$ \\
Qwen3.5 27B & $-1.136$ & $-1.187$ & $-1.268$ & $-1.337$ & $-1.142$ \\
Source draft & & & & & $-1.096$ \\
\midrule
Spearman & 1.000 & 1.000 & 0.976 & 0.976 & \\
\bottomrule
\end{tabular}}
\caption{Gate ablation. Equal-form standardized Bradley--Terry macro when an output
fails at $t$ or more violation tags ($t{\geq}1$ is the published gate), with no gate
(None), and with the source draft added as a ninth candidate to the ungated fit
(+Draft). Spearman is against the published order of the eight refiners.}
\label{tab:ablation}
\end{table}

\begin{table}[htbp]\centering\small\setlength{\tabcolsep}{4pt}
\begin{tabular}{@{}lrrr@{}}
\toprule
Model & Audio & Story & Essay \\
\midrule
GPT-5.5 & 0.999 & 0.978 & 0.986 \\
Claude Opus 4-6 & 0.997 & 0.980 & 0.997 \\
DeepSeek V4-Pro & 0.951 & 0.748 & 0.911 \\
Gemini 3.1 Pro & 0.829 & 0.910 & 0.929 \\
Grok 4.3 & 0.911 & 0.783 & 0.860 \\
Gemma 4 26B & 0.883 & 0.779 & 0.761 \\
Sarvam 105B & 0.625 & 0.520 & 0.554 \\
Qwen3.5 27B & 0.458 & 0.555 & 0.576 \\
\midrule
Pairs & 9{,}600 & 4{,}000 & 2{,}800 \\
\bottomrule
\end{tabular}
\caption{Each refiner's prose-only win rate against its own source draft (ties count
one half), over all refinements, gate-passing or not.}
\label{tab:drafts}
\end{table}

\FloatBarrier

\section{Faithfulness failure families}\label{app:failfamily}
\begin{figure}[htbp]\centering
\includegraphics[width=0.9\linewidth]{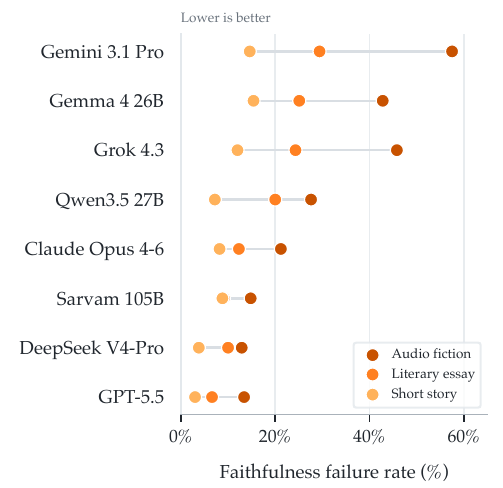}
\caption{Faithfulness-failure rate by model and form. Each model carries three points,
one per prose form; the rate varies both across models and, within a model, across
forms (for example Gemini 3.1 Pro fails far more on audio fiction than on essays or
stories). Lower is better.}
\label{fig:failrate}
\end{figure}

\begin{figure*}[htbp]\centering
\includegraphics[width=0.98\textwidth]{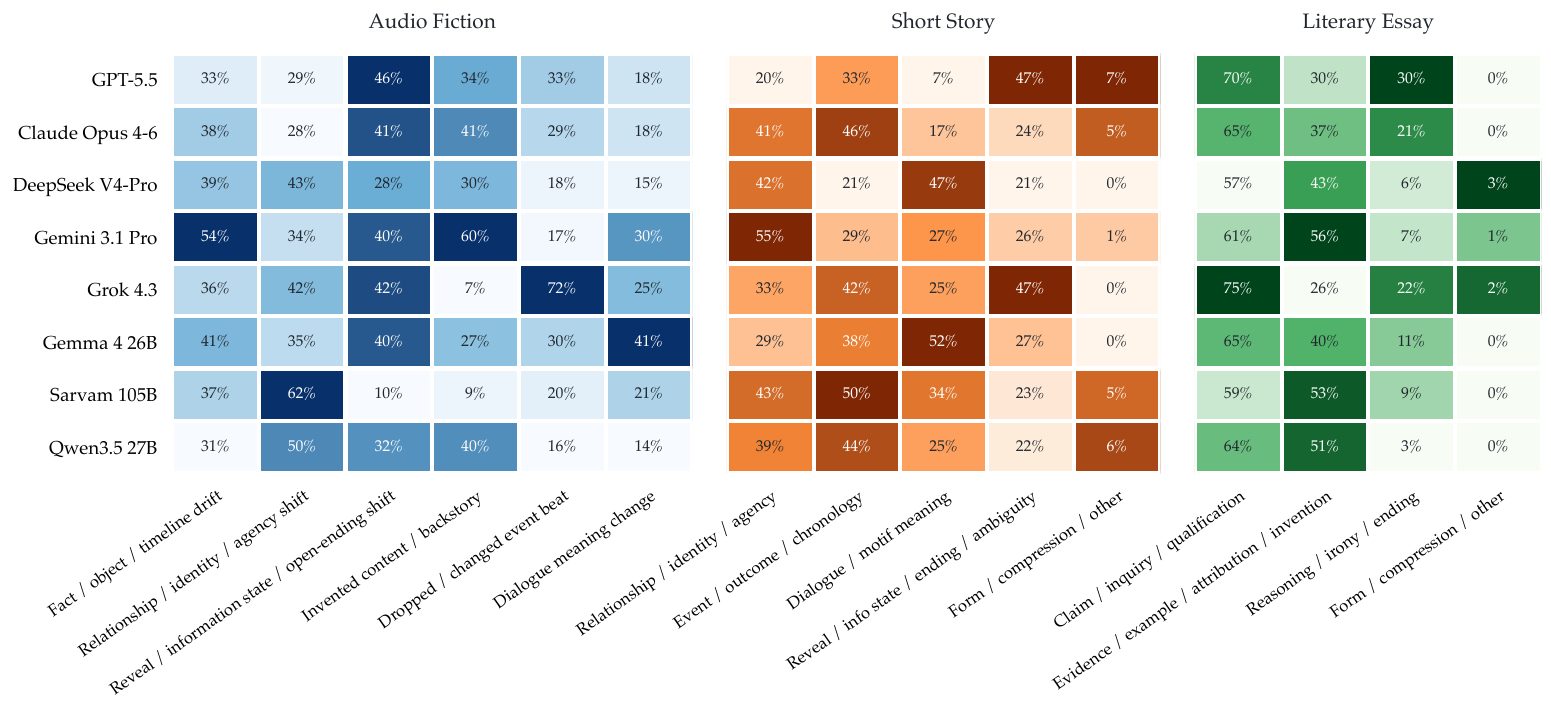}
\caption{Failure-family composition by model and form. Three panels (audio fiction,
short stories, literary essays); within each panel rows are the eight refiners by
leaderboard rank and columns are that form's failure families, each cell the fraction of
the model's gate-failed outputs carrying a tag in that family. Failures are multi-label,
so rows do not sum to one.}
\label{fig:failfamilymodel}
\end{figure*}

Figure~\ref{fig:failrate} gives per-model failure rates, Table~\ref{tab:failfamily} gives
the complete form-specific failure-family distributions, and Figure~\ref{fig:failfamilymodel}
breaks these families down by model within each form. Failures are multi-label, so shares
do not sum to one.

\begin{table}[htbp]\centering\small
\begin{tabular}{@{}l p{4.6cm} r r@{}}
\toprule
Form & Failure family & Count & Share \\
\midrule
Audio & Fact, object, or timeline drift & 1{,}160 & 40.9\% \\
Audio & Relationship, identity, or agency shift & 1{,}103 & 38.9\% \\
Audio & Reveal or information-state shift & 1{,}058 & 37.3\% \\
Audio & Invented content or backstory & 939 & 33.1\% \\
Audio & Dropped or changed event beat & 911 & 32.2\% \\
Audio & Dialogue-meaning change & 741 & 26.2\% \\
\midrule
Story & Relationship, identity, or agency & 143 & 39.2\% \\
Story & Event, outcome, or chronology & 141 & 38.6\% \\
Story & Dialogue or motif meaning & 116 & 31.8\% \\
Story & Reveal, information state, or ending & 107 & 29.3\% \\
Story & Form, compression, or other & 8 & 2.2\% \\
\midrule
Essay & Claim, inquiry, or qualification & 312 & 65.1\% \\
Essay & Evidence, example, or attribution & 206 & 43.0\% \\
Essay & Reasoning, irony, or ending & 59 & 12.3\% \\
Essay & Form, compression, or other & 4 & 0.8\% \\
\bottomrule
\end{tabular}
\caption{Complete form-specific failure-family distributions (shares over each form's
failed outputs). The audio corpus contains 2{,}833 gate-failed outputs, the short-story
corpus 365, and the essay corpus 479.}
\label{tab:failfamily}
\end{table}
\FloatBarrier

\FloatBarrier

\section{Generator-family kinship}\label{app:kinship}
Four refiners share a family sibling in the held-out generator pool. Recomputing each
form's Bradley--Terry leaderboard after removing every sibling-matched comparison
preserves the rank order exactly in all three forms: audio fiction drops 4{,}144 of
33{,}600 outcomes, short stories 1{,}813 of 14{,}000, and essays 1{,}330 of 9{,}800,
each with Spearman $\rho=1.0$ against the headline. The per-refiner sibling versus
non-sibling both-pass win-rate splits are reported in Table~\ref{tab:kinship}.

\begin{table}[htbp]
\centering\small
\setlength{\tabcolsep}{3.5pt}
\begin{tabular}{@{}lrrr@{}}
\toprule
Refiner & \shortstack{Sibling\\WR} & \shortstack{Non-sibling\\WR} & $\Delta$ \\
\midrule
\multicolumn{4}{@{}l}{\textit{Audio fiction}} \\
GPT-5.5 & 0.964 & 0.976 & -0.012 \\
Claude Opus 4-6 & 0.872 & 0.834 & +0.037 \\
Gemini 3.1 Pro & 0.639 & 0.667 & -0.027 \\
DeepSeek V4-Pro & 0.470 & 0.466 & +0.004 \\
\addlinespace[2pt]
\multicolumn{4}{@{}l}{\textit{Short stories}} \\
GPT-5.5 & 0.8612 & 0.8927 & -0.0316 \\
Claude Opus 4-6 & 0.8796 & 0.8639 & +0.0158 \\
Gemini 3.1 Pro & 0.5628 & 0.6640 & -0.1012 \\
DeepSeek V4-Pro & 0.4255 & 0.4391 & -0.0136 \\
\addlinespace[2pt]
\multicolumn{4}{@{}l}{\textit{Literary essays}} \\
GPT-5.5 & 0.7838 & 0.8879 & -0.1041 \\
Claude Opus 4-6 & 0.8566 & 0.8610 & -0.0044 \\
Gemini 3.1 Pro & 0.6431 & 0.6920 & -0.0489 \\
DeepSeek V4-Pro & 0.5469 & 0.5462 & +0.0007 \\
\bottomrule
\end{tabular}
\caption{Generator-family kinship audit. For the four refiners sharing a model family
with a held-out draft generator, entries are both-pass empirical win rates on
sibling-family and non-sibling source drafts. $\Delta=\mathrm{WR}_{\mathrm{sib}}-
\mathrm{WR}_{\mathrm{non}}$; ties count as 0.5, and positive values favor
sibling-generated drafts.}
\label{tab:kinship}
\end{table}

\FloatBarrier

\section{Judge-family robustness}\label{app:dims}
For audio fiction, two independent judge families re-judged the same 4{,}997 gate-passing
GPT-involving headline comparisons (the both-pass prose comparisons), each under the
frozen headline rubric with both presentation orders collapsed. Claude Opus 4-6 preserves
the full rank order (Spearman $\rho=1.0$) and agrees with the primary judge on 0.902 of
collapsed outcomes; Kimi-K3, a separate model family evaluated at temperature 0,
likewise preserves the full order (Spearman $\rho=1.0$) with GPT-5.5 first, agreeing on
0.866 (4{,}326/4{,}997; 95\% Wilson CI [0.856, 0.875]). Substituting either family's
judgments and refitting leaves GPT-5.5 at rank 1, with the gate-decided comparisons
unchanged. A uniform leniency offset of about 0.2 separates the primary judge from the
auditors on the 1--5 dimensions, so only the ordering is comparable; and on a
deterministic re-gated audio sample the primary gate passes 0.856 of the top model's
outputs where the Claude Opus 4-6 auditor passes 0.981 ($n=160$; on the other models'
outputs, 0.714 against 0.971, $n=140$), so it is the stricter, with no sign of
own-family leniency. Human labels agree: over the 600-case gate audit, labellers
overturned 2 of the 37 GPT-5.5 outputs the gate passed (0.054), below the audit-wide
false-pass rate of 0.076.
The short-story and essay leaderboards rely on the single primary judge; extending the
independent-family audits to them is future work.

\FloatBarrier

\section{Human validation detail}\label{app:human}
\textbf{Readers.} All readers and labellers are native Hindi speakers, paid members of
our organization who gave informed consent and judged blind to model identity. Six
readers labelled audio-fiction preference, five of them on a common set of 28
comparisons; three readers formed each written preference panel, sharing 20
comparisons per form; three labellers produced the second round of the written gate
audit; and four annotators produced the 40 inherited audio gate labels. Individual
annotations, reader identities, and demographics are withheld.

\textbf{Gate audit.} Table~\ref{tab:calib} reports the pooled results. The gate audit
has 200 cases per form, 600 in all: the submitted 200 audio-fiction cases, unchanged, and
for each written form 50 first-round cases plus 150 second-round cases labelled under the
same blind rubric and stratified by model and gate verdict. Agreement is 0.930 for audio
fiction, 0.935 (187/200) for short stories, and 0.910 (182/200) for essays. The gate's
false-fail and false-pass rates are conditioned differently across the audits and are
therefore not pooled: for audio they are 0.073 (7 of 96 human-pass cases) and 0.067 (7 of
104 human-fail cases); the written audits are balanced around the automatic gate label,
so for short stories they are 0.074 (7 of 94 automatic fails) and 0.057 (6 of 106
automatic passes), and for essays 0.079 (8 of 101) and 0.101 (10 of 99).
Table~\ref{tab:gatemodel} gives agreement per refiner, where every interval overlaps the
others; on GPT-5.5, the judge and top model, labellers overturned 2 of its 37 gate
passes (\S\ref{sec:results}). Table~\ref{tab:gatefamily} gives written agreement by the
gate's failure family, kept form-conditioned as in \S\ref{sec:results}; it ranges from
0.879 to 0.973, and on essays whose claim, inquiry, or qualification changed it is 0.950
(57/60).

\textbf{Preference.} The Task~A preference rows pool exact numerators across
independently run panels only to summarize agreement. Essay preference is difficult for
both humans and the automatic judge, and the judge's lower essay agreement should be
read against the correspondingly low human-human agreement. Fleiss' $\kappa$ is reported
per panel and never pooled across independently annotated panels; each value rests on a
small shared set, while each leaderboard is fit on thousands of pairs.

\begin{table}[htbp]\centering\small
\setlength{\tabcolsep}{3pt}
\begin{tabular}{@{}p{4.0cm} r r r@{}}
\toprule
Metric & Value & 95\% CI & n \\
\midrule
Gate vs human (pooled) & 0.925 & [0.901, 0.943] & 600 \\
Headline judge vs individual & 0.526 & [0.484, 0.568] & 540 \\
Inter-human pairwise & 0.472 & [0.424, 0.521] & 400 \\
Judge vs held-out majority & 0.524 & [0.403, 0.642] & 63 \\
Human vs leave-one-out\newline majority & 0.529 & [0.451, 0.605] & 157 \\
\midrule
Fleiss' $\kappa$, audio (5 readers) & 0.030 & & 28 \\
Fleiss' $\kappa$, story (3 readers) & 0.099 & & 20 \\
Fleiss' $\kappa$, essay (3 readers) & 0.012 & & 20 \\
\bottomrule
\end{tabular}
\caption{Pooled human calibration and gate validation across the three forms. These are
judge-audit statistics, not corpus-level pass-rate estimates. Fleiss' $\kappa$ is per
panel, with $n$ the shared comparisons every panel reader labelled.}
\label{tab:calib}
\end{table}

\begin{table}[htbp]\centering\small
\setlength{\tabcolsep}{3pt}
\begin{tabular}{@{}lrrr@{}}
\toprule
Refiner & Agree & n & Agreement [95\% CI] \\
\midrule
GPT-5.5 & 61 & 66 & 0.924 [0.835, 0.967] \\
Claude Opus 4-6 & 61 & 65 & 0.938 [0.852, 0.976] \\
DeepSeek V4-Pro & 60 & 64 & 0.938 [0.850, 0.975] \\
Gemini 3.1 Pro & 81 & 90 & 0.900 [0.821, 0.947] \\
Grok 4.3 & 62 & 71 & 0.873 [0.776, 0.932] \\
Gemma 4 26B & 64 & 70 & 0.914 [0.825, 0.960] \\
Sarvam 105B & 64 & 66 & 0.970 [0.896, 0.992] \\
Qwen3.5 27B & 64 & 68 & 0.941 [0.858, 0.977] \\
\midrule
All refiners & 517 & 560 & 0.923 [0.898, 0.943] \\
\bottomrule
\end{tabular}
\caption{Gate agreement with human labels per refiner over the 400 written and 160
on-roster audio audit cases. The remaining 40 audio cases come from two construction-era
systems outside the leaderboard.}
\label{tab:gatemodel}
\end{table}

\begin{table}[htbp]\centering\small
\setlength{\tabcolsep}{3pt}
\resizebox{\columnwidth}{!}{%
\begin{tabular}{@{}llrr@{}}
\toprule
Form & Failure family & n & Agreement [95\% CI] \\
\midrule
Story & Event, outcome, chronology & 44 & 0.909 [0.788, 0.964] \\
Story & Relationship, identity, agency & 40 & 0.925 [0.801, 0.974] \\
Story & Reveal, information, ending & 37 & 0.973 [0.862, 0.995] \\
Story & Dialogue or motif meaning & 33 & 0.879 [0.727, 0.952] \\
Essay & Claim, inquiry, qualification & 60 & 0.950 [0.863, 0.983] \\
Essay & Evidence, example, attribution & 47 & 0.894 [0.774, 0.954] \\
Essay & Reasoning, irony, ending & 25 & 0.920 [0.750, 0.978] \\
Both & Form, compression, other & 3 & 3 of 3 \\
\bottomrule
\end{tabular}}
\caption{Written gate agreement by failure family over the 400-case written audit,
counted on cases where the gate assigned a tag in that family. Tags are multi-label, so a
case can count in more than one row.}
\label{tab:gatefamily}
\end{table}

\FloatBarrier

\section{Prompt stack and disclosure}\label{app:prompts}
Table~\ref{tab:prompts} lists the nine prompts, their shared core, and the
form-conditioned content. The prompts share one construction, gate, and preference
architecture across the three forms, using explicit form cards where prose-form
expectations differ; the complete per-form prompt files are in the release.
The construction filter, gate, and headline-judge prompts (03, 05, and 06) are summarized
below in form-general terms that state both the narrative and essay conditioning; the
per-form specifics are itemized in Table~\ref{tab:prompts}.

\begin{table*}[htbp]\centering\small
\begin{tabular}{@{}p{2.6cm} p{5.6cm} p{7.4cm}@{}}
\toprule
Prompt & Shared core & Form-conditioned content \\
\midrule
01 Outline & Source to style-neutral outline & Source-form schema and metadata \\
02 Draft & Outline to draft $S$ & Audio, short-story, or essay form card \\
03 $S$-vs-$G$ filter & Accept draft $S$ & Form-specific acceptability criteria \\
04 Refinement & Preserve content, improve Hindi & Performability, narrative, or essay craft \\
05 Gate & Binary $R$-vs-$S$ eligibility & Narrative tags vs essay claim/evidence tags \\
06 Headline judge & Holistic A/B/Tie preference & Form-specific editorial rubric \\
07 J1 & Idiomatic Hindi, precision, flow & Register and translationese by form \\
08 J2 & Structure and progression & Dramatization or argumentative composition \\
09 J3 & Voice and engagement & Dialogue voice or authorial voice \\
\bottomrule
\end{tabular}
\caption{The nine prompts: shared construction-gate-preference architecture with explicit
form cards where prose-form expectations differ.}
\label{tab:prompts}
\end{table*}

\begin{promptbox}{Prompt 03: Source-versus-gold construction filter (build time only)}
\small\sffamily
\textbf{What it decides.} Whether a candidate draft $S$ keeps the core content of its
golden source $G$, checked before $G$ is detached. It judges content only, never prose
quality, and plays no part in leaderboard judging.\\[2pt]
\textbf{Audio fiction.} Pass if the draft keeps the main events, characters,
relationships, reveals, dialogue meanings, and ending state, however plain its prose.
Fail for a dropped or changed major event, a changed relationship or reveal, changed
dialogue meaning, invented major content, a changed ending or cliffhanger, a draft too
compressed to verify, or explicit or unsafe content.\\[2pt]
\textbf{Short stories and essays.} Pass by default, including local omissions, ordinary
compression, plainer register, and reordering. Fail only when a change alters a core
anchor of the work: for stories, the central situation or conflict, a decisive event or
causal turn, a driving identity or relationship, a central reveal, agency that drives the
outcome, or the ending; for essays, the central inquiry or stance, a principal claim, an
essential reasoning move, evidence the position depends on, or the final position. Also
fail a wrong prose form or a substantial English or Romanized continuation.\\[2pt]
\textbf{Output.} A pass or fail verdict with fixed reason tags and a short note; the
written versions also return paired evidence spans and repair instructions, used when
the draft is regenerated.
\end{promptbox}

\begin{promptbox}{Prompt 05: Faithfulness gate (binary, runs before scoring)}
\small\sffamily
\textbf{What it decides.} A content-preservation check between the source draft $S$ and a
refinement $R$ that judges source content only, never prose quality, and sets eligibility
for the leaderboard.\\[2pt]
\textbf{Narrative forms.} Fail if $R$ drops an important beat; changes an event, outcome,
or ending; alters a fact, number, or timeline; changes a plot-bearing object; changes who
acted or why; alters what a character says, knows, or decides; changes an identity or
relationship; adds or removes a reveal or shifts the information state; resolves or
introduces ambiguity; or invents new content.\\[2pt]
\textbf{Essays.} Fail if $R$ changes, drops, or adds a claim; alters or invents evidence,
examples, or attribution; changes a qualification or the strength of a claim; or changes
the line of reasoning, irony, or conclusion.\\[2pt]
\textbf{Decision rule.} Does the change alter what the source establishes, or only how
vividly the same thing is expressed? If the former, fail. The judge returns a pass or fail
verdict with a short evidence span and, on failure, one or more fixed content-violation
tags.
\end{promptbox}

\begin{promptbox}{Prompt 06: Headline pairwise judge (holistic A/B/Tie, prose-only)}
\small\sffamily
\textbf{What ``better'' means.} Reading two versions of the same unit as written prose in
its form, the stronger one is the version an editor would rather release: natural, native,
register-correct Hindi; drama, argument, or reflection that carries the reader; and
form-appropriate voice. The judge sees no source or outline and weighs the writing, not
the content.\\[2pt]
\textbf{Limits.} Effect must be carried in natural Hindi; a version whose effect rides on
English calque or translationese reads as imported and wins no points for it, and ornate
vocabulary that hurts readability is no advantage.\\[2pt]
\textbf{Output.} A single holistic verdict, A, B, or Tie, with a one-line reason. The
harness runs both presentation orders and collapses them, and the collapsed outcomes feed
the per-form Bradley--Terry fit.
\end{promptbox}

\FloatBarrier

\section{Diagnostic lexicon}\label{app:diag}
Human reviewers compiled and filtered the candidate phrasings into a frozen registry
before scoring. The formulaic-phrase registry holds 272 reviewed surfaces for audio fiction and 274 for
the written forms; the two additional written-form surfaces form a rhetorical-template
family. The calque registry holds 120 reviewed surfaces, shared across all three forms.
Matching is exact substring matching over lightly normalized Devanagari: FPB counts
distinct matched surfaces per 1{,}000 word-like tokens (mean of per-output rates), and
calque is the share of outputs with at least one match. Exact-match calque detection
can miss calques phrased in other words \citep{rahman-etal-2026-calques}, so these calque rates likely
undercount calquing. The two additional written-form
surfaces are essay rhetorical templates that do not apply to audio fiction, so the audio
FPB uses the 272-surface registry (Table~\ref{tab:lexical}). The complete lexicons, with
attested examples per surface, are in the release; representative calques
appear in Table~\ref{tab:calques}. Tables~\ref{tab:fpbexamples} and~\ref{tab:driftexamples}
give further qualitative examples: representative
formulaic-phrase families with the recurring phrase highlighted, and human-confirmed
faithfulness-drift minimal pairs caught by the gate.

\begin{table*}[htbp]\centering\small
\setlength{\tabcolsep}{5pt}
\begin{minipage}[t]{0.46\textwidth}\centering
\textbf{Panel A. Formulaic-phrase burden (per 1{,}000 words)}\\[2pt]
\begin{tabular}{@{}lrrr@{}}
\toprule
Model & Audio & Story & Essay \\
\midrule
GPT-5.5 & 3.39 & 1.05 & 0.05 \\
Claude Opus 4-6 & 4.42 & 1.45 & 0.04 \\
DeepSeek V4-Pro & 3.54 & 0.87 & 0.04 \\
Gemini 3.1 Pro & 3.15 & 1.04 & 0.06 \\
Gemma 4 26B & 4.28 & 0.87 & 0.03 \\
Sarvam 105B & 2.78 & 0.71 & 0.04 \\
Grok 4.3 & 2.73 & 0.66 & 0.07 \\
Qwen3.5 27B & 3.51 & 0.80 & 0.05 \\
\bottomrule
\end{tabular}
\end{minipage}\hfill
\begin{minipage}[t]{0.46\textwidth}\centering
\textbf{Panel B. Calque output-hit rate}\\[2pt]
\begin{tabular}{@{}lrrr@{}}
\toprule
Model & Audio & Story & Essay \\
\midrule
GPT-5.5 & 77.3\% & 13.6\% & 2.9\% \\
Claude Opus 4-6 & 81.2\% & 18.2\% & 1.1\% \\
DeepSeek V4-Pro & 70.8\% & 15.2\% & 2.0\% \\
Gemini 3.1 Pro & 82.1\% & 25.4\% & 2.9\% \\
Gemma 4 26B & 77.5\% & 24.2\% & 1.4\% \\
Sarvam 105B & 59.3\% & 13.2\% & 0.9\% \\
Grok 4.3 & 43.5\% & 11.6\% & 0.9\% \\
Qwen3.5 27B & 76.3\% & 15.0\% & 1.1\% \\
\bottomrule
\end{tabular}
\end{minipage}
\caption{Cross-form lexical diagnostics. Panel~A: formulaic-phrase burden per 1{,}000
word-like tokens (mean of per-output distinct-surface rates). Panel~B: percentage of
outputs with at least one of the 120 reviewed calque surfaces; all three forms use the
identical calque list. These are descriptive diagnostics, not ranking
inputs. The formulaic registry is 272 surfaces for audio fiction and 274 for the written
forms; the two additional surfaces are written-form rhetorical templates that do not
apply to audio.}
\label{tab:lexical}
\end{table*}

\begin{table*}[htbp]\centering\small
\begin{tabular}{@{}p{2.6cm} p{13cm}@{}}
\toprule
Phrase family & Example refinement (recurring phrase highlighted) \\
\midrule
Dialogue / voice tags & \hi{उसकी \textcolor{cFpb}{\textbf{आवाज़ काँपी}}, "मैं जानता हूँ वह कौन है।"} \\
Dialogue / voice tags & \hi{"मुझे मूर्ख मत समझ," परांजय ने \textcolor{cFpb}{\textbf{धीमे लेकिन साफ़ लहज़े में कहा}}।} \\
Body-reaction beats & \hi{उसने \textcolor{cFpb}{\textbf{दाँत भींचे}} और झटके से खड़ा हो गया।} \\
Body-reaction beats & \hi{उसकी \textcolor{cFpb}{\textbf{उँगलियाँ काँप रही थीं}}।} \\
Sensory / atmosphere padding & \hi{...अपने पीछे एक \textcolor{cFpb}{\textbf{भारी और दमघोंटू सन्नाटा}} छोड़ते हुए।} \\
Sensory / atmosphere padding & \hi{\textcolor{cFpb}{\textbf{शहर की रोशनियाँ पीछे छूटती}} जा रही थीं...} \\
Silence / tension filler & \hi{कोठरी के भीतर \textcolor{cFpb}{\textbf{सन्नाटा इतना गहरा था कि}} लगता था जैसे मौत खुद सांसें रोककर बैठी हो।} \\
Silence / tension filler & \hi{और उसी पल कमरे में एक नई, \textcolor{cFpb}{\textbf{भारी खामोशी छा गई}}।} \\
Generic emotional telling & \hi{उस आदमी की आवाज़ में अब \textcolor{cFpb}{\textbf{घबराहट साफ़ थी}}।} \\
Generic emotional telling & \hi{...लेकिन उनकी आवाज़ में \textcolor{cFpb}{\textbf{वह बेचैनी थी जो}} सालों की जुदाई के बाद स्वाभाविक थी।} \\
\bottomrule
\end{tabular}
\caption{Formulaic-phrase burden: representative stock phrase families recurring above
the human baseline, with the recurring phrase highlighted (real model
outputs). Quantified per model in Figure~\ref{fig:signature}; a descriptive signature,
not a ranking input.}
\label{tab:fpbexamples}
\end{table*}

\begin{table*}[htbp]\centering\small
\begin{tabular}{@{}p{5.8cm} p{5.8cm} p{3.6cm}@{}}
\toprule
Source phrase (S) & Rewritten phrase (R) & Faithfulness issue \\
\midrule
\hi{...कल सुबह तक वापस \textcolor{cOk}{\textbf{नहीं}} आ पाऊँगा} & \hi{...कल सुबह तक \textcolor{cDrift}{\textbf{ही}} वापस आ पाऊँगा} & Negation flipped. \\
\hi{अभी \textcolor{cOk}{\textbf{सैर}} करो} & \hi{अभी \textcolor{cDrift}{\textbf{आराम}} करो} & Instruction changed from walking to resting. \\
\hi{उसने भिजवाया \textcolor{cOk}{\textbf{है}}} & \hi{उसने ही भिजवाया \textcolor{cDrift}{\textbf{होगा}}} & Fact changed to speculation. \\
\hi{अपनी अगली चाल सोच रही थी} & \hi{\textcolor{cDrift}{\textbf{शायद}} अपनी अगली चाल सोच रही थी} & Stated fact changed to possibility. \\
\hi{कपूर खानदान का वारिस} & \hi{कपूर खानदान के \textcolor{cDrift}{\textbf{इकलौते}} वारिस} & Invented ``sole heir'' detail. \\
\hi{मेरा हुक्म \textcolor{cOk}{\textbf{ज़्यादा}} बड़ा है} & \hi{मेरा हुक्म भी \textcolor{cDrift}{\textbf{उतना ही}} बड़ा है} & Greater authority changed to equal authority. \\
\hi{गाँव का सबसे \textcolor{cOk}{\textbf{बुज़ुर्ग}}} & \hi{जीवन के \textcolor{cDrift}{\textbf{सौ-बाइस साल}} देख चुके} & Invented specific age. \\
\hi{संदीप (कंट्रोल का ज़िम्मेदार)} & \hi{वही संदीप \textcolor{cDrift}{\textbf{जिसके साथ कितनी उड़ानें भरी थीं}}} & Invented shared history. \\
\hi{(वह सूरज को चाहती थी)} & \hi{ज़िद उसके भीतर \textcolor{cDrift}{\textbf{बरसों से}} जलती रही थी} & Invented years-long backstory. \\
\hi{कान पकड़कर \textcolor{cOk}{\textbf{उठक-बैठक}}} & \hi{इधर-उधर \textcolor{cDrift}{\textbf{छटपटाने}} लगा, जैसे व्यायाम} & Specific punishment changed to vague action. \\
\bottomrule
\end{tabular}
\caption{Faithfulness drift caught by the gate (human-confirmed minimal pairs): the
source span in \textcolor{cOk}{green}, the altered rewrite span in
\textcolor{cDrift}{red}. Drift appears across the roster, including the top-ranked
models, which is why the gate runs on every output. All text is model-generated.}
\label{tab:driftexamples}
\end{table*}
\begin{center}
\begin{minipage}{\columnwidth}
\centering\small
\begin{tabular}{@{}l p{3.8cm} r r@{}}
\toprule
Form & Cluster & Sources & Share \\
\midrule
Audio & Fantasy or cultivation & 335 & 27.9\% \\
Audio & Drama, family, or social & 293 & 24.4\% \\
Audio & Thriller or suspense & 195 & 16.2\% \\
Audio & Romance & 133 & 11.1\% \\
Audio & Crime or mystery & 119 & 9.9\% \\
Audio & Supernatural, historical, or other & 125 & 10.4\% \\
\midrule
Story & General literary short story & 358 & 71.6\% \\
Story & Fable or folk story & 117 & 23.4\% \\
Story & Children's story & 23 & 4.6\% \\
Story & Experimental story & 2 & 0.4\% \\
\midrule
Essay & Cultural reflective & 209 & 59.7\% \\
Essay & Literary reflective & 111 & 31.7\% \\
Essay & Personal reflective & 17 & 4.9\% \\
Essay & Literary critical & 9 & 2.6\% \\
Essay & Argumentative or expository & 4 & 1.1\% \\
\bottomrule
\end{tabular}
\captionof{table}{Source diversity by prose form. Percentages are within each prose form. The
audio-fiction clusters are genre families; the audio tail row groups supernatural,
historical, action, science-fiction, and mythological items.}
\label{tab:diversity}
\end{minipage}
\end{center}

\FloatBarrier

\section{Source diversity and reproducibility}\label{app:repro}
\textbf{Sources.} The audio-fiction sources are professionally edited episodes from a
private contemporary corpus; the short stories are drawn from Hindi Wikisource, the
BHAAV corpus \citep{kumar2019bhaav}, and Hindi Discourse
\citep{dhanwal2020discourse}, and the literary essays from Hindi Wikisource. Rights
and release are handled at the item level and some material is restricted (Ethics
Statement).

\textbf{Construction filter.} Prompt 03 (Appendix~\ref{app:prompts}) compares each
candidate draft with its source $G$ before $G$ is detached, with form-conditioned
wording. Counting candidate drafts that also pass the hard length and script check, it
rejected 307 of 1{,}612 audio-fiction drafts (19.0\%), 124 of 624 short-story drafts
(19.9\%), and 44 of 394 essay drafts (11.2\%). A rejected written draft is regenerated by
a different generator from the held-out pool, which 13.8\% of short-story and 12.0\% of
essay sources needed; a source whose audio draft is rejected is replaced by another
episode. The audio rejections carry multi-label reason tags: a major event changed
(38\% of rejections), a character or relationship changed (37\%), the ending or
cliffhanger changed (24\%), a major event missing (24\%), dialogue meaning changed
(21\%), and a new event or reveal added (17\%). The filter is automatic and has no human
audit of its own.

\textbf{Lengths.} Lengths are counted in word-like tokens; model-tokenizer counts are
omitted because Devanagari tokenization differs across the eight systems. Drafts average
1{,}468 tokens for audio fiction, 1{,}186 for short stories, and 1{,}067 for essays, and
Table~\ref{tab:lengths} gives each refiner's output length relative to its draft. Against
$G$, the mean per-item ratio of draft to source length is 0.657 for short stories and
0.664 for essays; the ratio of corpus totals is lower (0.490 and 0.461) because the draft
length is bounded, so long sources are compressed proportionally more.

\begin{table}[htbp]\centering\small\setlength{\tabcolsep}{4pt}
\begin{tabular}{@{}lrrr@{}}
\toprule
Refiner & Audio & Story & Essay \\
\midrule
GPT-5.5 & 1.598 & 1.156 & 1.099 \\
Claude Opus 4-6 & 1.119 & 0.963 & 0.987 \\
DeepSeek V4-Pro & 1.102 & 1.017 & 1.026 \\
Gemini 3.1 Pro & 1.661 & 1.109 & 1.117 \\
Grok 4.3 & 0.775 & 0.839 & 0.909 \\
Gemma 4 26B & 1.169 & 1.046 & 1.039 \\
Sarvam 105B & 1.003 & 0.955 & 1.009 \\
Qwen3.5 27B & 1.142 & 1.007 & 1.014 \\
\midrule
Mean draft (tokens) & 1{,}468 & 1{,}186 & 1{,}067 \\
\bottomrule
\end{tabular}
\caption{Mean refinement length as a multiple of the mean draft length, per refiner and
form, in word-like tokens.}
\label{tab:lengths}
\end{table}

\noindent\textbf{Settings.} The benchmark uses one fixed, documented setting per model
and stage; Table~\ref{tab:temperatures} summarizes the temperature settings.

\begin{center}
\begin{minipage}{0.95\columnwidth}
\centering
\small
\setlength{\tabcolsep}{7pt}
\begin{tabular}{@{}lr@{}}
\toprule
Stage & Temperature \\
\midrule
Outline extraction & 0.1 \\
Draft generation and refinement & 0.65 \\
Construction filter & 0.0 \\
Gate and evaluation judges & 0.0 where supported \\
\bottomrule
\end{tabular}
\captionof{table}{Temperature settings by stage. GPT-5.5 adapters omit unsupported
temperature fields.}
\label{tab:temperatures}
\end{minipage}
\end{center}

\textbf{Model documentation.} Primary technical reports, model cards, or release announcements document the construction models Gemini 3.1 Flash-Lite
\citep{google2026gemini31flashlite}, GPT-5.4 Mini \citep{openai2026gpt54mini},
Claude Sonnet 4.5 \citep{anthropic2025sonnet45}, Gemini 3 Flash
\citep{google2025gemini3flash}, Kimi K2 \citep{moonshot2025kimik2}, GLM-5
\citep{glm5team2026}, MiniMax M2.7 \citep{minimax2026m27}, and Mistral Medium
3.5 \citep{mistral2026medium35}; and the evaluated systems GPT-5.5
\citep{openai2026gpt55}, Claude Opus 4-6 \citep{anthropic2026opus46}, Gemini
3.1 Pro \citep{google2026gemini31pro}, DeepSeek V4 \citep{deepseek2026v4},
Gemma 4 \citep{google2026gemma4}, Sarvam 105B \citep{sarvam2026sovereign},
Qwen3.5 27B \citep{qwen2026qwen35}, Grok 4.3 \citep{xai2026grok43}, and
Kimi-K3 \citep{moonshot2026kimik3}. The Kimi K2 citation documents the model
family used by the requested K2.6 route.

\textbf{Diversity.} The 1{,}200 audio-fiction sources span 82 shows (median 14 episodes
per show) across roughly eleven genre families: fantasy and cultivation 27.9\%, family and
social drama 24.4\%, thriller and suspense 16.2\%, romance 11.1\%, crime and mystery 9.9\%,
with smaller supernatural, historical, and mythological tails. 70.8\% of items are
contemporary, and episodes average 4.1 scenes and 7.3 characters; all such metadata is
approximate, extracted at the outline step. Table~\ref{tab:diversity} gives that breakdown
alongside the written-form source diversity: the 500 short stories and 350 essays each
spread across several sub-types, with percentages within each form. Scene- and
character-count histograms and the full tables are in the release.

\end{document}